\pdfoutput=1

\documentclass[11pt]{article}

\usepackage[preprint]{acl}

\usepackage{times}
\usepackage{latexsym}

\usepackage[T1]{fontenc}

\usepackage[utf8]{inputenc}

\usepackage{microtype}

\usepackage{inconsolata}

\usepackage{url}
\usepackage{hyperref}

\usepackage{microtype}
\usepackage{amsmath}
\usepackage{amsthm}
\usepackage{graphicx}

\usepackage{algorithm}
\usepackage{algorithmicx}
\usepackage[noend]{algpseudocode}

\usepackage{xcolor}
\usepackage{inconsolata}
\usepackage{multirow}
\usepackage{booktabs}
\usepackage{pifont}
\usepackage{bbding} 
\usepackage{array}
\usepackage{graphicx}
\usepackage{booktabs}
\usepackage{subcaption}

\usepackage{listings}
\usepackage{enumitem} 

\usepackage{array}
\usepackage{makecell}

\title{FlowBench: Revisiting and Benchmarking Workflow-Guided\\ Planning for LLM-based Agents}

\author {
    Ruixuan Xiao$^{1}$\thanks{~ Equally contribution. Work done while Ruixuan Xiao was interning at Alibaba Group.}, Wentao Ma$^{2}$\footnotemark[1], Ke Wang$^2$, {Yuchuan Wu}$^2$\\
      \textbf{Junbo Zhao$^1$,} \textbf{Haobo Wang$^1$}\thanks{~ Corresponding authors.}, \textbf{Fei Huang$^{2}$}, \textbf{Yongbin Li$^2$}\footnotemark[2]\\
    $^1$Zhejiang University \quad $^2$Alibaba Group\\
    \texttt{\{xiaoruixuan,j.zhao,wanghaobo\}@zju.edu.cn}\\
    \texttt{\{mawentao.mwt,wk258730,shengxiu.wyc,f.huang,shuide.lyb\}@alibaba-inc.com}
}

\begin{document}
\maketitle
\begin{abstract}
LLM-based agents have emerged as promising tools, which are crafted to fulfill complex tasks by iterative planning and action.
However, these agents are susceptible to undesired planning hallucinations 
when lacking specific knowledge for expertise-intensive tasks.
To address this, preliminary attempts are made to enhance planning reliability by incorporating external workflow-related knowledge. Despite the promise, such infused knowledge is mostly disorganized and diverse in formats, lacking rigorous formalization and comprehensive comparisons.
Motivated by this, we formalize different formats of workflow knowledge and present \textbf{FlowBench}, the first benchmark for workflow-guided planning. FlowBench covers 51 different scenarios from 6 domains, with knowledge presented in diverse formats.
To assess different LLMs on FlowBench, we design a multi-tiered evaluation framework. We evaluate the efficacy of workflow knowledge across multiple formats, and the results indicate that current LLM agents need considerable improvements for satisfactory planning. We hope that our challenging benchmark can pave the way for future agent planning research. {Our benchmark and scripts will be released shortly.}

\end{abstract}

\section{Introduction}
The impressive advances of large language models (LLMs) \cite{DBLP:journals/corr/abs-2302-13971,DBLP:conf/iclr/ZengLDWL0YXZXTM23,DBLP:journals/corr/abs-2303-08774} have spurred the evolution of LLM-driven agents \cite{DBLP:journals/fcsc/WangMFZYZCTCLZWW24,DBLP:journals/corr/abs-2308-00352} for complex task solving across diverse domains~\cite{DBLP:journals/corr/abs-2307-07924,DBLP:journals/corr/abs-2308-00352,DBLP:journals/corr/abs-2307-13854}. While recent works have made strides in advancing tool utilization abilities of agents through prompt construction \cite{DBLP:conf/iclr/YaoZYDSN023,DBLP:conf/nips/YaoYZS00N23} and multi-agent collaboration \cite{DBLP:journals/corr/abs-2308-00352,DBLP:journals/corr/abs-2309-17288,DBLP:journals/corr/abs-2307-07924}, the restricted scope of LLMs' intrinsic parametric knowledge can lead to the undesired phenomenon of planning hallucinations~\cite{DBLP:journals/corr/abs-2403-03101} -- LLMs can exhibit uncontrollable actions that conflict with task knowledge and such adverse effects obstruct their practical application in knowledge-intensive tasks.

\begin{figure}[!t]
     \centering
         \includegraphics[width=\columnwidth]{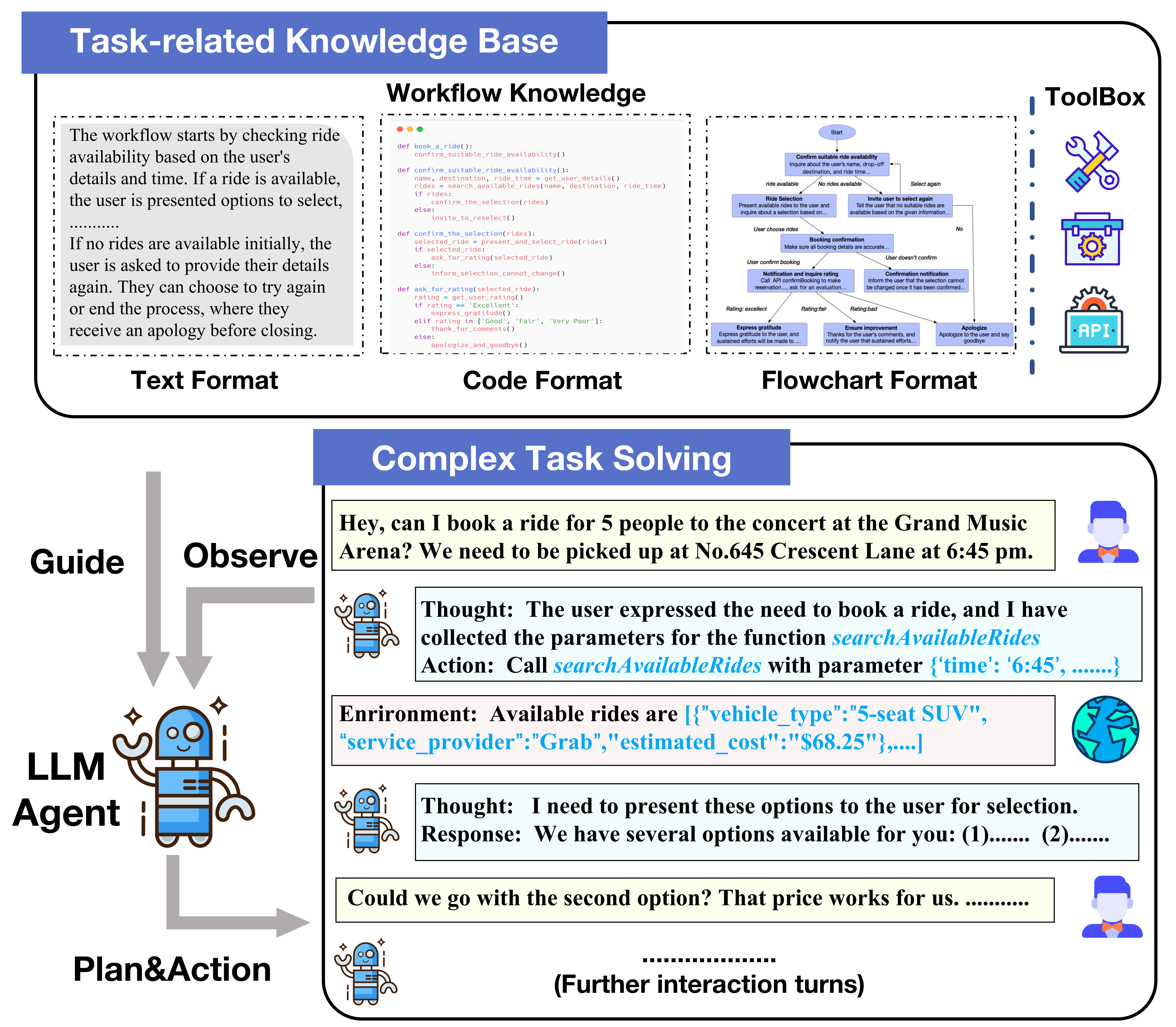}
     \caption{The procedure of workflow-guided agent planning. The agent is provided with workflow knowledge in various formats and prompted to plan the next action.}
     \label{fig:head}
\end{figure}

On the other hand, mitigating hallucinations by leveraging external knowledge is an ongoing key topic for LLM research \cite{DBLP:journals/corr/abs-2312-10997,DBLP:conf/www/BaekCCHJ24,DBLP:journals/corr/abs-2301-12652,DBLP:conf/www/BaekCCHJ24}. As for LLM-agent planning, some preliminary efforts have been made to enhance the 
planning quality of agents by incorporating workflow-related knowledge \cite{DBLP:journals/corr/abs-2402-11163,DBLP:conf/kcap/ChhikaraZIFM23,DBLP:journals/corr/abs-2311-10751}. For example, KnowAgent~\cite{DBLP:journals/corr/abs-2403-03101} employs the explicit action knowledge summarized in natural language format to regulate the planning trajectory. ProAgent~\cite{DBLP:journals/corr/abs-2311-10751} enhances the efficiency of complex tasks through control flows described in Python code. 
Despite the promise, these nascent works commonly shape workflow-related knowledge in miscellaneous and unmethodical formats, lacking in-depth exploration and rigorous benchmarking comparisons concerning the role of workflows. Hence, it still remains underexplored how to formalize, utilize and evaluate such workflow-related knowledge for LLM-based agents across different real-world scenarios.

To this end, we meticulously explore different formats of workflow knowledge and establish a systematic benchmark for evaluating workflow-guided agent planning. 
Our study initially revisits and formalizes various embedding formats of workflow knowledge, including natural language, symbolic code, and flowchart schema, as depicted in Figure~\ref{fig:head}. To perform a comparative evaluation, we present \textbf{FlowBench}, the first comprehensive benchmark for workflow-guided agent planning. FlowBench covers an extensive taxonomy (6 domains, 22 roles, 51 scenarios) and different knowledge formats (text, code, flowchart) to synchronize with real-world applications comprehensively. The benchmark data is constructed through a three-phase pipeline of task collection, workflow organization, and session generation. FlowBench features numerous distinct characteristics, such as coverage, difficulty, expert-level annotation, and support for multi-round user-agent interaction, as summarized in Table~\ref{tab:other_benchmarks}.

To conduct a reliable assessment, we design a holistic evaluation framework that comprises two distinct granularities: (i)-Static turn-level evaluation that focuses on single-step planning.  
(ii)-Dynamical session-level evaluation that simulates sequential planning. Based on such evaluation mechanisms, we evaluate LLMs across varying levels of capability and embedded with different formats of workflow knowledge. 
We observe that even the best-performing LLM, GPT-4o~\cite{HelloGPT4o}, struggles to deliver satisfactory performance in certain tasks (43.2\% and 40.9\% success rate). By contrasting various workflow formats, we discover that flowcharts struck the best trade-off among {performance}, {adaptability}, and user-friendliness.
All these results illuminate the profound challenge that FlowBench offers to the present LLM-agent designs and steer the focus of future exploration.

Our contributions are summarized as follows,
\begin{itemize}
    \item {As far as we know, we are the first to revisit and formalize the definition of different represented workflow knowledge systematically. 
    }
    \item {
     We construct FlowBench, the first comprehensive benchmark for workflow-guided agent planning, which covers 51 scenarios across 6 domains, with different workflow formats. 
     }
    \item {
    We present a holistic evaluation framework to achieve reliable comparisons. We provide a comparative analysis of various LLMs with different workflow formats, indicating promising directions for agent planning.
    }
\end{itemize}

\begin{table*}[ht]
 \tabcolsep=0.25cm
\small
\centering
\begin{tabular}{l|ccc|cc}
\toprule
\multirow{2}{*}{\textbf{Benchmark}} 
& 
\multirow{2}{*}{\textbf{Domain}} 
& 
\multirow{2}{*}{\textbf{Purpose\&Task}} 
& 
\multirow{2}{*}{%
  \begin{tabular}{@{}c@{}}
    \textbf{Workflow} \\
    \textbf{Knowledge}
  \end{tabular}%
}
& 
\multicolumn{2}{c}{%
\textbf{Multi-turn Interaction}
}
\\
&&&&\textit{Environment} & \textit{User} 
\\
\midrule

\ PlanBench & Logistics, Blocksworld & Textual Script Generation & \color{purple}{\XSolidBrush} & \color{purple}{\XSolidBrush} & \color{purple}{\XSolidBrush}  \\

\ TravelAgent & Traveling & Travel Plan Generation & \color{purple}{\XSolidBrush} & \color{purple}{\XSolidBrush} & \color{purple}{\XSolidBrush}  \\

\ KnowAgent & QA, Text Games & Online Task Planning & Text & \color{teal}{\Checkmark} & \color{purple}{\XSolidBrush} \\

\ ProAgent & Robotic Process Automation & Online Task Planning & Code & \color{teal}{\Checkmark} & \color{purple}{\XSolidBrush} \\ 

 \midrule

\multirow{2}{*}{
\textbf{FlowBench} 
}
& 
\multirow{2}{*}{
\begin{tabular}{@{}c@{}}
    6 Domains, 22 Roles, \\
    51 Scenarios
\end{tabular}%
 
}
& 
\multirow{2}{*}{
Online Task Planning
}
& 
\multirow{2}{*}{
\begin{tabular}{@{}c@{}}
Text, Code, \\
Flowchart
\end{tabular}%
}
& 
\multirow{2}{*}{
\color{teal}{\Checkmark}
}
& 
\multirow{2}{*}{
\color{teal}{\Checkmark}
}
\\
& & & & &
\\
\bottomrule
\end{tabular}
\caption{Comparison of FlowBench with some related studies and benchmarks, including ProAgent~\cite{DBLP:journals/corr/abs-2311-10751}, PlanBench~\cite{DBLP:conf/nips/ValmeekamMHSK23}, TravelAgent~\cite{DBLP:journals/corr/abs-2402-01622} and KnowAgent~\cite{DBLP:journals/corr/abs-2403-03101}.
}
\label{tab:other_benchmarks}
\end{table*}
\section{Related Work}
\subsection{LLM-Driven Agents}
The evolving abilities of LLMs have fostered explorations for LLM-driven agents \cite{DBLP:journals/fcsc/WangMFZYZCTCLZWW24,DBLP:journals/corr/abs-2308-00352,DBLP:journals/corr/abs-2309-17288}, which are designed for complex task solving and have shown great potential in various fields, such as software development \cite{DBLP:journals/corr/abs-2307-07924,DBLP:journals/corr/abs-2308-00352,DBLP:journals/corr/abs-2309-17288}, web navigation \cite{DBLP:conf/nips/DengGZCSWSS23,DBLP:journals/corr/abs-2307-13854}, and healthcare support \cite{DBLP:conf/aaai/YangZZZXJZ24,zhang-etal-2023-huatuogpt}. Early approaches strive to improve the consecutive reasoning ability of LLMs, where LLMs are prompted to engage in progressive and comprehensive thinking, such as Chain of Thought (CoT) \cite{DBLP:conf/nips/Wei0SBIXCLZ22}, ReAct \cite{DBLP:conf/iclr/YaoZYDSN023}, and Tree of Thought \cite{DBLP:conf/nips/YaoYZS00N23}. 
The latter strand of studies focuses on enhancing the tool utilization \cite{DBLP:conf/nips/SchickDDRLHZCS23,DBLP:journals/corr/abs-2307-16789,DBLP:journals/corr/abs-2305-15334} and multi-agent collaboration \cite{DBLP:journals/corr/abs-2308-00352,DBLP:journals/corr/abs-2309-17288,DBLP:journals/corr/abs-2307-07924} of LLMs.  

\subsection{Knowledge-Augmented Agents}
Knowledge-augmented LLMs \cite{DBLP:journals/corr/abs-2312-10997,DBLP:conf/www/BaekCCHJ24,DBLP:journals/corr/abs-2301-12652,DBLP:conf/www/BaekCCHJ24} have been a prevailing topic, as leveraging external knowledge can effectively improve response factuality \cite{DBLP:journals/corr/abs-2311-07914,DBLP:journals/corr/abs-2404-00942,DBLP:conf/nips/LiuZHCTLWYGZL23}.
As for LLM-based agents, task-related knowledge and expertise are also important for making correct plans and decisions. To achieve this, nascent attempts \cite{DBLP:journals/corr/abs-2402-11163,DBLP:conf/kcap/ChhikaraZIFM23} have been made to enhance the reliability of planning in designated tasks by incorporating workflow-related knowledge, such as action rules described in natural language \cite{DBLP:conf/icml/HuangAPM22,DBLP:journals/corr/abs-2403-03101}, domain knowledge represented as knowledge graphs \cite{DBLP:journals/corr/abs-2404-14741,DBLP:journals/corr/abs-2402-11163}, and control flow expressed in code \cite{DBLP:journals/corr/abs-2401-00812,DBLP:journals/corr/abs-2311-10751}. However, these works mostly craft workflow knowledge in diverse and sloppy representation formats, and there is a lack of formalized consensus and comprehensive benchmarking comparisons concerning the effect of infused workflow knowledge. 

\subsection{Evaluation of LLM Agents}
To assess the abilities of LLMs as agents, a plethora of benchmarks have been established \cite{DBLP:journals/corr/abs-2308-03688,DBLP:journals/corr/abs-2401-13178,DBLP:journals/corr/abs-2310-01557}. These works are mostly tailored for evaluating generic abilities like tool utilization \cite{DBLP:journals/corr/abs-2307-16789,chen2024teval}, code generation \cite{DBLP:journals/corr/abs-2308-03688,DBLP:journals/corr/abs-2401-13178} and embodied interaction \cite{DBLP:journals/corr/abs-2310-01557}. Regarding workflow-integrated planning, it remains questionable how to systematically assess the agent's comprehension and utilization ability of the integrated workflow knowledge.

\begin{figure*}[!t]
     \centering
         \includegraphics[width=0.95\linewidth]{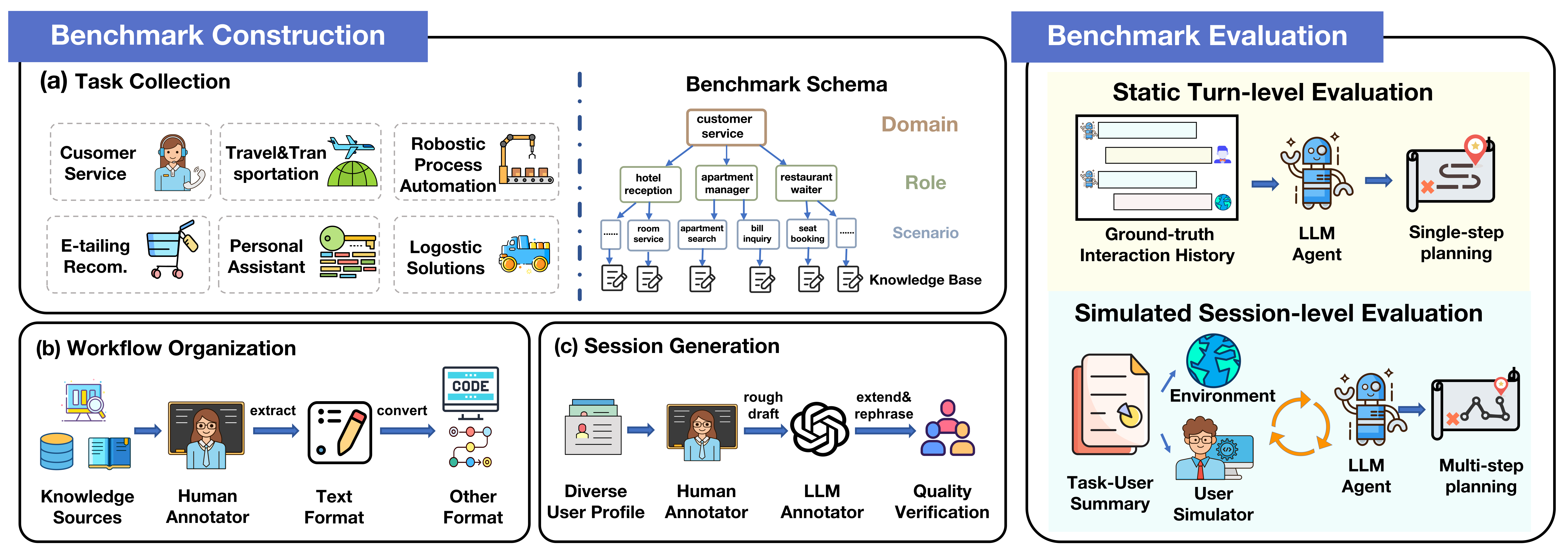}
     \caption{{Overview of FlowBench. Our benchmark schema is structured in a top-down multi-level hierarchy (domain - role - scenario - knowledge). The benchmark construction process on the left contains three phases (a,b,c). The evaluation framework on the right encapsulates static turn-level and simulated session-level assessment.}}
     \label{fig:overview}
\end{figure*}

\section{Workflow Formalization}

\subsection{Task Formulation}\label{sec:task_formulation}
In our research scope, we consider an LLM-driven agent $\mathcal{M}_{\theta}$ tackling a specific downstream task. This agent $\mathcal{M}_{\theta}$ is equipped with externel task-specific knowledge base $\mathcal{B}=\{\mathcal{K},\mathcal{P}\}$, which contains workflow-related knowledge $\mathcal{K}$ and toolbox $\mathcal{P}$.  In most existing agent studies, agents $\mathcal{M}_{\theta}$ are initially provided with complete and detailed task descriptions as initial input. Nevertheless, such a non-interactive assumption is not realistic in real-world multi-round online planning. 
To this end, we generalize this to a more realistic setting of multi-turn interactions, allowing users to incrementally refine and modify their requirements through ongoing conversations with the agent.

Formally, assuming that at the iteration step $i$ during task-solving, the agent $\mathcal{M}_{\theta}$ resides in state $s_i$. Upon receiving feedback $e_i$ from the external environment and conversational input $u_i$ from the user, the agent is expected to harness both its built-in parametric commonsense $\theta$ and external workflow knowledge base $\mathcal{B}$ to map out an action plan $a_{i+1}$ and provide a conversational reply $r_{i+1}$ to the user. With this action, the agent shifts to the subsequent state $s_{i+1}$,
\begin{equation}\label{formula:definition}
\{a_{i+1},s_{i+1},r_{r+1}\}\leftarrow \mathcal{M}_{\theta}({\mathcal{H}_i},\mathcal{B})
\end{equation}
In this formula, $\mathcal{H}_i$ refers to the interaction history encapsulating all the entries $\{u_t,e_t,s_t,a_t,r_t\}_{t=0}^{i}$ from iteration $0$ to $i$, which includes both current input $\{u_i,e_i\}$ from user/environment and the historical information $\{s_i,a_i,r_i\}$ from all past steps, as shown in Figure~\ref{fig:head}. 
After $T$ turns of such sequential planning, we expect the agent $\mathcal{M}_{\theta}$ to successfully accomplish the user's tasks by formulating a correct action plan trajectory $\{a_0,a_1,...,a_T\}$ and a satisfactory response $\{r_0,r_1,...,r_T\}$.

\subsection{Revisiting Different Workflow Formats}
\label{sec:formats}
Workflow-related knowledge is a set of representations of pipeline-related facts, which refers to the understanding of how processes or tasks are structured and executed within a specific context, such as a business or project environment. 
The content of different workflow knowledge primarily falls into the following categories:
(i)-\textbf{Operation process related}, which describes the required steps and their sequential order to complete tasks.
(ii)-\textbf{Condition/rule related}, which delineates the actions to be taken when certain conditions or parameters are met.
(iii)-\textbf{Tool/data related}, which contains the utilization techniques for different tools and the mechanisms of data processing.

As mentioned earlier, workflow-related knowledge can be embedded in a broad spectrum of formats~\cite{DBLP:journals/corr/abs-2403-03101,DBLP:journals/corr/abs-2311-10751}. Thus, we begin by revisiting different workflow manifestations. Based on the level of abstraction, we primarily analyze three distinct representation forms: 
\begin{itemize}
    \item {
    \textbf{Text} format that is conveyed through vanilla natural language documentation.
    } 
    \item {
    \textbf{Code} format that follows programming language standards, e.g., Python pseudo-code. 
    }
    \item {
    \textbf{Flowchart} format that is expressed in a low-code visual-programming diagram syntax, e.g., Markdown Mermaid notation.
    
    }
\end{itemize}
These three forms of workflow knowledge are depicted in Figure~\ref{fig:head} (see Appendix~\ref{app:add_demo} for more examples). Their respective fortes and flaws can be broadly outlined with the following two criteria,

\begin{itemize}

    \item \textbf{Efficacy and Precision}: Considering knowledge expression efficiency, text format is flexible in expressing complex concepts, but it can lead to excessive token consumption and undesired semantic ambiguity. Conversely, codes and flowcharts leverage structured symbols to enhance their precision and efficacy.
    \item \textbf{User friendliness}: User-friendliness is primarily contingent upon how facilitative it is for users to edit and comprehend. Although text format allows for easy editing, it can be cumbersome to pinpoint specific knowledge within lengthy documents. For code, the demand for programming expertise to edit it also presents a challenge. By contrast, the flowchart presents a more user-friendly alternative, allowing users with no coding experience to edit easily and providing visualization interfaces to improve their understanding.
    
\end{itemize}

Overall, it can be concluded that natural language is more expressive but less efficient and often ambiguous, while programming code is more structured, concise, and precise, though less intuitive for user comprehension and editing.
Flowcharts create a bridge, integrating the powerful expressiveness of natural language with the high efficiency of symbolic code, and providing highly user-friendly options for visualization and editing.

\section{FlowBench}
The overview of its construction process and evaluation framework is displayed in Figure~\ref{fig:overview}. In what follows, we will elaborate on this in detail.

\subsection{Hierarchy of Benchmark Schema}\label{sec:schema}

 FlowBench consists of a collection of downstream \textbf{domains} (distribution shown in Section~\ref{sec:bench_con}). Each domain incorporates a range of agent \textbf{roles}. Each of these roles is further delineated by a suite of specific \textbf{scenarios} (i.e., fundamental task). Each scenario is related to the corresponding task-related \textbf{knowledge base}, which encompasses task background, workflow-related knowledge, and a collection of tools to be employed. 
 
 A comprehensive top-down example is depicted in the left part of Figure~\ref{fig:overview}. In the domain of customer services, several agent roles are configured, including restaurant waiter, hotel reception, and apartment manager. The role of apartment manager is then in charge of two scenarios: apartment search and bill inquiry. Each of these scenarios is then related to its task-related knowledge base.

\subsection{Benchmark Construction} 
\label{sec:bench_con}
The construction pipeline of FlowBench is structured into three phases: scenario collection, workflow organization, and session generation.

\paragraph{Task Collection}
For a benchmark to effectively assess workflow-guided planning ability, its diversity and wide coverage are essential. To achieve this, we initially draw inspiration from the collected tasks in existing works \cite{DBLP:journals/corr/abs-2010-11853} and further conduct extensive extensions. Our task collection is targeted towards both personal customers and business enterprises. Consequently, the roles and scenarios of FlowBench are meticulously collected from the following 6 domains:

\begin{table}[!t]
    \centering
    \small
    \renewcommand\arraystretch{1.1}
    \tabcolsep=0.045cm
    \begin{tabular}{l|cccc}
        \toprule
        \textbf{Domain} & \textbf{Role} & \textbf{Scenario}
      & \textbf{Session} & \textbf{Turn} \\
        \midrule
    {Customer service} &   {4}  & {12}  & {114} & {1167} \\
    {Personal assistant} &   {3}  & {7}  & {92} & {821} \\
    {E-tailing Recommandation} &   {2}  & {5}  & {32} & {330} \\
    {Travel\&Transportation} &   {4}  & {9}  & {135} & {1421} \\  
    {Logistics solutions} &   {3}  & {6}  & {61} & {521} \\
     {Robotic process automation} &   {6}  & {12}  & {102} & {1053} \\
    \midrule
    {Overall} & 22 & 51 & 536 & 5313  \\
        \bottomrule
    \end{tabular}
    \caption{Domain statistics of FlowBench. }
    \label{tab:domain_sta}
\end{table}

\begin{itemize}[itemsep=1pt, parsep=1pt]
     \item {\textbf{Customer Service} provides advice, reception, reservation, and after-sales support. 
    }

    \item {\textbf{Personal Assistant} is to provide personal management and solution services.
    }
    
     \item {\textbf{E-tailing Recommendation} aids in product discovery and purchase during shopping.
     }

     \item {\textbf{Travel\&Transportation} offers support for travel and transportation arrangements.
     }

     \item {\textbf{Logistics Solutions} assist in the management of express delivery and logistics services. 
     }
     
      \item {\textbf{Robotic Process Automation (RPA)} focuses on automating and streamlining complex work processes,   especially in business contexts.
      }

\end{itemize}

As mentioned in benchmark schema (Section~\ref{sec:schema}), each of these domains contains several agent roles (22 in total), and each role in these domains further includes several task scenarios (51 in total). The complete scenarios and roles for each domain can be found in the Appendix and the domain distribution is also drawn in Figure~\ref{fig:overview}.

\paragraph{Workflow Organization}
After collecting task scenarios, next we describe the extraction and structuring of workflow-related knowledge for each scenario. Instead of brainstorming, we consult some existing references, such as professional knowledge corpora (e.g., WiKiHow~\cite{DBLP:journals/corr/abs-1810-09305}), websites of workflow knowledge (e.g., Zapier~\cite{zapier}), and search engine results. 

Integrating information from these sources, we initially summarize the workflow-related expertise point by point into a natural language document. Then different human annotators are required to independently verify the correctness, completeness, and non-redundancy of these knowledge documents. Following this, we employ GPT-4~\cite{openai2023gpt4} to convert such text format knowledge into code and flowchart formats following the specified standards and incorporate manual verification to ensure knowledge consistency. An example of workflow knowledge embedded in different formats is provided in Appendix~\ref{app:add_demo}.  After structuring the workflow knowledge, we proceed to organize the tool invocation information involved in the workflow. Adhering to the GPT-4 function calling format, we outline the description for each tool call, the parameters for inputs/outputs, and the corresponding information for each parameter.

\begin{table*}[ht]

    \renewcommand{\arraystretch}{1.1}
    \tabcolsep=0.18cm
    \small
    \centering
    \begin{tabular}{lc|ccccc|ccccc}
    \toprule
    \multirow{3}{*}{\raisebox{-1\height}{\textbf{Model}}} & \multirow{3}{*}{\raisebox{-1\height}{\textbf{Format}}} & \multicolumn{5}{c|}{\textbf{\textsc{Single-Scenario}}} & \multicolumn{5}{c}{\textbf{\textsc{Cross-Scenario}}} \\ 
    \addlinespace[2pt]
    \cline{3-12}
    \addlinespace[2pt]
    & & \multicolumn{3}{c}{Tool Invocation}  & \multirow{2}{*}{\begin{tabular}[c]{@{}c@{}} Parameter \\ F1 \end{tabular}}& \multirow{2}{*}{\begin{tabular}[c]{@{}c@{}}Response \\ Score \end{tabular}}  
    & \multicolumn{3}{c}{Tool Invocation}  & \multirow{2}{*}{\begin{tabular}[c]{@{}c@{}}Parameter \\ F1 \end{tabular}} & \multirow{2}{*}{\begin{tabular}[c]{@{}c@{}}Response \\ Score \end{tabular}}\\
    \addlinespace[1pt]
    & & P & R & F1 & & & P & R & F1 
    \\
    \midrule
 \multirow{4}{*}{GPT-4o} & None & 73.8 & 60.3 & 66.3  & 76.5  & 7.80 & 60.8 & 51.2 & 55.4 & 70.8 & 7.85  \\
  & Text & {75.9} & 63.6 & 69.1 & 77.2  & 8.22 & 66.0 & 54.9 & 59.9 & 71.0 &  \textbf{8.34} \\
 & Code & 73.7 & 64.9 & 69.0 & 78.3  & 8.30 & 63.2 & 53.9 &  58.2 & {70.7} &  8.23 \\
 & Flowchart & \textbf{83.4} & \textbf{68.9} & \textbf{75.5} & \textbf{80.9} & \textbf{8.38} &  \textbf{73.9} & \textbf{56.0} & \textbf{63.7} & \textbf{72.2}  & 8.29  \\  \midrule

 \multirow{4}{*}{GPT-4-Turbo} & None & 68.3 & 52.1 & 59.1 & 71.7 & 7.55 & 53.9 & 45.6 & 49.2 & 63.9 & 7.77  \\
  & Text & {70.2} & 56.8 & 62.9 & 74.5 & 7.94 & 55.3 & 47.2 & 51.1 & 64.3 &  8.05 \\
 & Code & 73.7 & 61.4 & 67.0 & 78.8 & 7.98 & 53.7 & 49.4 & 51.4 & 65.1 &  8.03 \\
 & Flowchart & \textbf{78.9} & \textbf{69.3} & \textbf{73.6}  & \textbf{79.2}  & \textbf{8.06} & \textbf{67.2} & \textbf{58.1} & \textbf{62.3} & \textbf{71.5} & \textbf{8.10}  \\ \midrule

\multirow{4}{*}{GPT-3.5-Turbo} & None & {58.5} & 51.9 & 55.0 & 67.9 &  7.01 & 52.5 & 42.7 & 47.2 & 66.2 & 7.19 \\
& Text & 63.4 & 56.8 & 59.8 & 72.5 & 7.28 & 53.5 & 48.9 & 51.0 & 69.0 & 7.39  \\
 & Code & 60.7 & 55.0 & 57.9 & 69.2  & 7.33 & 53.1 & 47.7 & 50.2 & 70.2 & \textbf{7.33}  \\
 & Flowchart & \textbf{71.6} & \textbf{59.6} & \textbf{65.4} & \textbf{76.3}  & \textbf{7.39} & \textbf{60.9} & \textbf{51.0} & \textbf{55.6}  & \textbf{70.8} & 7.30  \\

    \bottomrule
\end{tabular}
\caption{{Performance comparisons of different LLMs equipped with different formats of workflow knowledge under \textbf{static turn-level} evaluation. \textbf{Bold} entries indicate superior results.}}
\label{tab:main_results_static}
\end{table*}

\paragraph{Interactive Session Generation}
With the organized workflow-related knowledge at hand, our subsequent endeavor is to synthesize the ground-truth user-agent interactive sessions for every scenario. Such generated sessions will serve as the prerequisite for our evaluation framework, whose details are discussed later in Section~\ref{section:eval_frame}. Our principle is to enhance the diversity and authenticity of generated sessions. To achieve this, we prompt GPT-4 to generate diverse user profiles under each scenario to enhance diversity, which includes \textit{user background}, \textit{user targets}, \textit{response tones} (i.e., response style).  For data authenticity, we further intentionally incorporate some out-of-scenario items, such as casual chit-chat and irrelevant off-scope intents, into the user profiles, mimicking real-world scenarios.

Given generated user profiles, a collaborative annotation strategy is employed, which assigns a cooperative pair of a human annotator and an LLM annotator (GPT-4) for both the user and agent side. Specifically, for each interaction turn, on the user side, the human annotator crafts the subsequent user intent and dialogue draft based on task context and interaction history, which is then verified and rephrased by the LLM annotator to match the required response tone. As for the agent side, the human annotator leverages the workflow knowledge to formulate the next action plan and craft an initial response draft, which is subsequently reviewed and embellished by the LLM annotator.  This collaborative generation process continues iteratively until the entire session is fully annotated.

\paragraph{Data Verification}
To ensure data quality, we incorporate human verification at each stage during benchmark construction. Three human annotators participate in the quality verification process. Each submission from an annotator will be subject to cross-verification by the remaining two annotators.

\begin{table*}[ht]

    \renewcommand{\arraystretch}{1.1}
    \tabcolsep=0.2cm
    \small
    \centering
    \begin{tabular}{lc|ccccc|ccccc}
    \toprule
    \multirow{3}{*}{\raisebox{-1\height}{\textbf{Model}}} & \multirow{3}{*}{\raisebox{-1\height}{\textbf{Format}}} & \multicolumn{5}{c|}{\textbf{\textsc{Single-Scenario}}} & \multicolumn{5}{c}{\textbf{\textsc{Cross-Scenario}}} \\ 
    \addlinespace[2pt]
    \cline{3-12}
    \addlinespace[2pt]
    & & \multicolumn{3}{c}{Tool Invocation}  & \multirow{2}{*}{\begin{tabular}[c]{@{}c@{}}Success \\ Rate\end{tabular}}& \multirow{2}{*}{\begin{tabular}[c]{@{}c@{}}Task \\ Progress\end{tabular}}  
    & \multicolumn{3}{c}{Tool Invocation}  & \multirow{2}{*}{\begin{tabular}[c]{@{}c@{}}Success \\ Rate\end{tabular}} & \multirow{2}{*}{\begin{tabular}[c]{@{}c@{}}Task \\ Progress\end{tabular}}\\
    \addlinespace[1pt]
    & & P & R & F1 & & & P & R & F1 
    \\
    \midrule
 \multirow{4}{*}{GPT-4o} & None & {67.8} & 70.5 & 69.0 & 35.1 & 77.5  & 45.1 & 53.9 & 49.0 & 34.2 & 75.7   \\
  & Text & 73.7 & 81.2 & 77.2 & 41.7 & \textbf{83.4}  & \textbf{50.5} & 58.7 & 54.1  & \textbf{40.9} & 81.2   \\
 & Code & {70.9} & 79.6 & 75.0 & \textbf{43.2} & {83.1}  & 47.7 & 56.9 & 51.9 & 38.2 & 77.9    \\
 & Flowchart & \textbf{74.3} & \textbf{82.6} & \textbf{78.3} & {42.7} & 82.2  & {50.2} & \textbf{60.4} & \textbf{54.7} & 40.1 & \textbf{81.3}  \\ \midrule

 \multirow{4}{*}{GPT-4-Turbo} & None & {62.9} & 64.0 & 63.5 &  31.9 & 76.7 & 43.6 & 47.0 & 45.2 & 31.1 & 75.3 \\
  & Text & 65.9 & 69.9 & 67.8 &  \textbf{41.5} & \textbf{83.8} & 47.6 & 49.3 & 48.4 & 36.8 & 79.6  \\
 & Code & \textbf{79.3} & 64.7 & 71.2  & 37.8 & 83.6 & 47.5 &\textbf{52.9} & 50.0 & 37.3 & \textbf{82.0}    \\
 & Flowchart & {70.5} & \textbf{78.3} & \textbf{74.2}  & 40.6 & {82.3}  & \textbf{50.1} & {51.2} & \textbf{50.6} & \textbf{39.4} & 81.1   
 \\ \midrule

\multirow{4}{*}{GPT-3.5-Turbo} & None & 37.6 & 50.5 & 43.1 & 29.8 & 63.7 & 24.7 & 30.7 & 27.4 & 22.4 & 61.1 \\
& Text & {44.1} & \textbf{55.6} & \textbf{49.1} &  \textbf{37.0} & \textbf{69.7} & \textbf{25.6} & 31.5 & 28.2 & \textbf{25.2} & 63.9    \\
 & Code & \textbf{44.5}  & 52.9 & 48.3 &  31.3 & 66.3 & 24.5 & 31.9 & 27.8 & 24.1 & 62.9   \\
 & Flowchart & 43.6 & 54.3 & 48.4 & {34.9} & {68.6}  & 24.9 & \textbf{33.0} & \textbf{28.5} & 24.7 & \textbf{67.0}  \\ 
 
    \bottomrule
\end{tabular}
\caption{{Performance comparisons of different LLMs equipped with different formats of workflow knowledge under \textbf{simulated session-level} evaluation. \textbf{Bold} entries indicate superior results.}}
\label{tab:main_results_simulated}
\end{table*}

\paragraph{Benchmark Statistics}
The statistics of FlowBench are shown in Table~\ref{tab:domain_sta}, and the comparisons with some related benchmarks are in Table~\ref{tab:other_benchmarks}. The construction cost is in Appendix~\ref{app:cons_cost}.

\subsection{Evaluation Framework}\label{section:eval_frame}

After benchmark construction, we propose a holistic and multi-faceted evaluation framework in this section. We categorize our evaluation scenarios according to task awareness, and present two distinct facets of static and simulated assessments.

\subsubsection{Task Awareness}
As previously mentioned in Section~\ref{sec:schema}, in our dataset schema, each agent role is paired with several respective task scenarios. Therefore, we divide our evaluation scenarios into single-scenario and cross-scenario evaluations based on whether the specified task scenario is known a priori. (i)-\textbf{Single-scenario} evaluation assumes a pre-determined task scenario. An agent role is provided with the workflow knowledge of that scenario and needs to navigate, plan, and execute actions within this single task scenario.
(ii)-\textbf{Cross-scenario} evaluation assumes that the specific scenario is unknown. The agent role is equipped with a versatile set of workflow knowledge covering all scenarios within the role scope. The agent needs to flexibly plan and switch between different scenario.
\subsubsection{Evaluation Protocols}\label{sec:eval_protocol}
As mentioned in Section~\ref{sec:task_formulation}, our setting of multi-turn user-agent interactions generalizes the traditional agent paradigm. Consequently, the conventional evaluation protocols are incompatible with our benchmark. To solve this, we design a holistic evaluation protocol at both turn and session level, as depicted in the right of Figure~\ref{fig:overview}.
\paragraph{Static Turn-level Evaluation.} 
Static turn-level evaluation operates based on the ground-truth sessions generated in Section~\ref{sec:bench_con}. Specifically, given a sampled ground-truth session $d$, for each turn $i$ in $d$, the agent is provided with the ground-truth interaction history of previous ($i-1$) turns from $d$, and then prompted to plan the current-turn action.
Next, we compare the predicted action $\tilde{a}_i$ and response $\tilde{r}_i$ with ground truth ${a}_i$,${r}_i$ for evaluation.

\paragraph{Simulated Session-level Evaluation.} 
To more comprehensively assess planning capabilities in real-world scenarios where actions are sequentially planned and carried out, a simulated session-level evaluation framework is proposed. 

Specifically, we construct a user simulator based on GPT-4. 
In order to ensure that the behavior of this simulated user aligns with humans, we also 
rely on the ground-truth sessions generated in Section~\ref{sec:bench_con}. 
We distill a task-user summary from each of these sessions via GPT-4, which concludes information of \textit{task background, user goals, tool invocation information}. Notably, this task-user summary is different from the generated user profile in Section~\ref{sec:bench_con}. The \textit{user goals} summarizes all the objectives of the user, while \textit{tool invocation information} outlines the expected parameters and values for tool usage. Based on this task-user summary, a predicted session is produced with the simulated user and the to-be-assessed agent for metric calculation. 
All the evaluation metrics for turn and session-level are discussed later in Section~\ref{sec:metric}.

\section{Experiment}

\subsection{Models and Baselines}\label{sec:metric}
\paragraph{LLMs.} 
Given the difficulty of solving complex tasks in practical scenarios, we mainly focus on examining the effectiveness of advanced LLMs. Hence we conduct experiments on several advanced LLMs with varying degrees of capability from OpenAI: GPT-4o~\cite{HelloGPT4o}, GPT-4-Turbo \cite{openai2023gpt4}, GPT-3.5-Turbo \cite{openai2022chatgpt}, 

\paragraph{Knowledge Formats.}
We explore performances with different embedded formats of workflow knowledge, including (i)-\textbf{Text} variant in natural language document formats, (ii)-\textbf{Code} variant that adheres to Python pseudo-code style, (iii)-\textbf{Flowchart} variant using low-code flowcharts following the markdown mermaid syntax. (iv)-\textbf{None} variant is further included, which only offers tool information, withholding the workflow-related knowledge. Notably text/code/flowchart are based on a consistent workflow and are merely different in formats.

\paragraph{Metrics.}
We evaluate turn and session-level performance based on step-wise planning accuracy and overall task completion.
\textbf{Tool invocation} is evaluated for both levels by precision (P), recall (R), and F1-score. An invocation is considered correct only when both tool name and all required parameters are correctly identified. On session level, we sequentially check each ground-truth invocation to verify if it matches any simulated one. 
On turn level, the F1-score of \textbf{parameter collection} is also included. We further evaluate \textbf{response quality} by preference scoring (full marks:10) based on correctness, helpfulness, and humanness via GPT-4. For session level, we evaluate the \textbf{success rates} and \textbf{task progress} via GPT-4.
(i)-Success rate describes the proportion of completely successful sessions. A session is deemed successful when all the goals specified in the task-user summary are achieved.
(ii)-Task progress indicates the percentage of goals completed within each session.
The evaluation prompts are in Appendix~\ref{app:prompt_eval}.

\paragraph{Implementation Details.}
During the inference, we adopt \texttt{gpt-4o-0513} version for GPT-4o, \texttt{gpt-4-0125-preview} version for GPT-4-Turbo, and \texttt{gpt-3.5-turbo-16k-0613} version for GPT-3.5-Turbo. In automatic evaluation for response preference, success rate, and average progress, we adopt \texttt{gpt-4-turbo-0125-preview} version for fair comparisons. The inference prompts utilize ReAct~\cite{DBLP:conf/iclr/YaoZYDSN023} framework consisting of thought, action, observation. More implementation details and examples are put in Appendix~\ref{app:more_imple}.

\begin{table}[t]

    \renewcommand{\arraystretch}{1.0}
    \tabcolsep=0.14cm
    \small
    \centering
    \begin{tabular}{l|ccc|ccc}
    \toprule
    \multirow{2}{*}{\textbf{Format}} & \multicolumn{3}{c|}{\textbf{\textsc{S-Scenario}}} & \multicolumn{3}{c}{\textbf{\textsc{C-Scenario}}} \\ 
    \addlinespace[2pt]
    \cline{2-7}
    \addlinespace[2pt]
    & {Tool} & {SR} & {TP} 
    & {Tool}  & {SR} & {TP}
    \\ %
    \midrule
{None w/o tool} &  53.0 & 28.9 & 72.1 & 31.7 & 25.3 & 71.3 \\ \midrule
{None} &  63.5 & 31.9 & 76.7 & 45.2 & 31.1 & 75.3 \\
{Text} &  67.8 & 41.5 & 83.8 & 48.4 & 36.8 & 79.6 \\
{Code} &  71.2 & 37.8 & 83.6 & 50.0 & 37.3 & \textbf{82.0} \\
{Flowchart} & 74.2 & 40.6 & 82.3 & 50.6 & 39.4 & 81.1 \\
\midrule
{Ensemble} &  \textbf{75.1} & \textbf{45.3} & \textbf{85.2} & \textbf{51.8} & \textbf{43.4} & {81.6} \\

\bottomrule

\end{tabular}
\caption{Performance for variant \textit{None w/o tool} and \textit{Ensemble} with GPT-4-Turbo under session-level evaluation. `Tool', `SR', and `TP' indicate F1 score of tool invocation, success rate and task progress respectively.
}
\label{tab:main_ablation}
\end{table}

\subsection{Main Results}
Table~\ref{tab:main_results_static} and ~\ref{tab:main_results_simulated} display the performance comparisons on the turn and session level. Based on these results, the following summarizations can be observed: 

\textbf{(i)}-When external workflow knowledge is absent and merely tool information is provided (variant \textit{None}), LLMs can still rely on their intrinsic commonsense to achieve a basic understanding of workflow. However, in some expertise-intensive domains, lacking workflow knowledge can result in notable degradation, as discussed in Section~\ref{sec:analysis}.

\textbf{(ii)}-The incorporation of workflow knowledge in different formats can significantly facilitate agent planning, achieving evident improvements at both turn level and session level. Cross-scenario transitions cause a discernible decline, yet workflow knowledge still provides a notable improvement. 

\textbf{(iii)}-The efficacy of different knowledge formats varies in different settings. The code format is less effective on weaker LLMs, potentially because the complex symbolic expressions impede information conveyance. The text format, on the other hand, continues to perform well on different LLMs. 

\textbf{(iv)}-Overall, the flowchart format generally produces the best performance. For static turn-level evaluations, it shows a notable advantage of 6.4\%, 6.5\% and 5.6\% for F1 score of tool invocation on different LLMs. For session level, the flowchart demonstrates comparable performance to other formats for task completion evaluation, while it still performs the best for tool invocation. We speculate that such a lead stems from its organized and comprehensible nature, which enables LLMs to conveniently pinpoint the current state for better planning. More analyses are put to Appendix~\ref{app:results}.

\begin{figure}[!t]
     \centering
     \begin{subfigure}{0.49\linewidth}
         \includegraphics[width=\columnwidth]{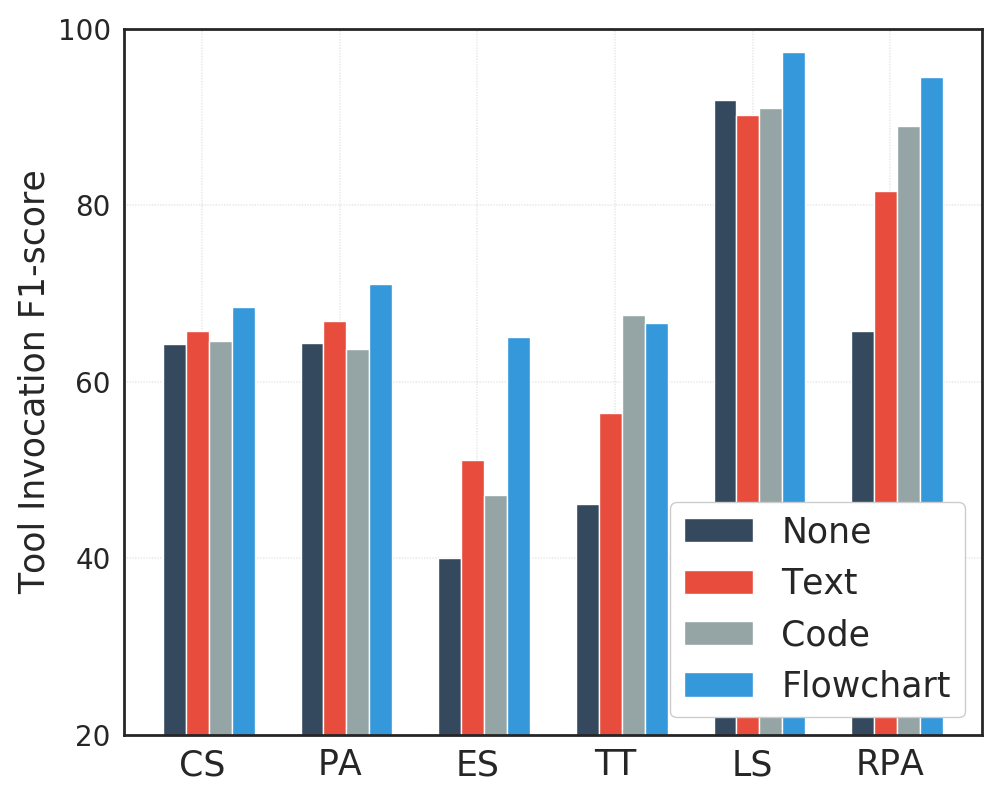}
         \caption{Single-Scenario.}
     \end{subfigure}
     \hfill
     \begin{subfigure}{0.49\linewidth}
         \includegraphics[width=\columnwidth]{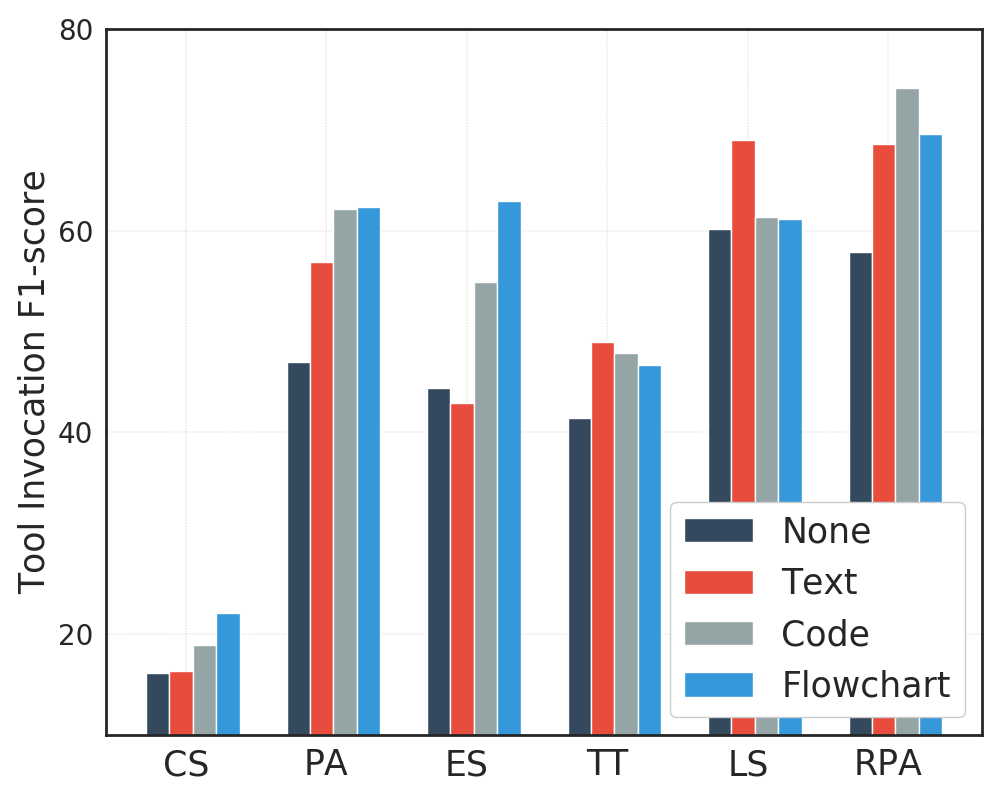}
         \caption{Cross-Scenario. }
     \end{subfigure}
     
     \caption{Domain-wise comparisons under session-level evaluation. Initials are adopted to denote domains. }
     \label{fig:perf_distri}
\end{figure}

\subsection{Further Analysis}\label{sec:analysis}
\paragraph{Domain-wise Performance}
We further investigate the performance distribution across domains. For experiments on GPT-4-Turbo, the domain-wise tool invocation F1-score is depicted in Figure~\ref{fig:perf_distri}. We can see that workflow knowledge plays a more significant role in those expertise-demanding domains such as RPA, E-tailing Recommendation, and Travel, greatly compensating for the lack of inherent task-specific expertise in LLMs.

\paragraph{Effect of Tool Information}
As mentioned in Section~\ref{sec:schema}, in our task-specific knowledge bases, alongside workflow-related knowledge that delineates work processes, there is also tool information describing the usage of different tools. 
Here, we further strip away the tool information from \textit{None} variant and investigate this \textit{None w/o tool} variant, from which descriptions of both tools and their parameters are completely eliminated (only tool and parameter names remain). As shown in Table~\ref{tab:main_ablation}, 
such a variant experiences further degradation, especially for cross-scenario settings.

\paragraph{Knowledge Formats Combination.}
We further explored the impact of combining different formats of workflow knowledge. As shown in Table~\ref{tab:main_ablation}, such a \textit{Ensemble} variant is provided with text, code and flowchart formats simultaneously. It shows additional performance gains, which demonstrate that different formats of consistent knowledge can complement each other and get combined to boost the agent's comprehension of workflows.

\section{Conclusions}
We introduce FlowBench, the first benchmark tailored for evaluating workflow-guided agent planning. We first revisit and formalize different workflow knowledge formats. Next, through extensive experiments on FlowBench, we find that flowchart formats achieve the best trade-off in terms of performance and user experience. The results further indicate that even the best-performing model, GPT-4o, fails to deliver satisfying results on challenging FlowBench.
We hope that our work can provide meaningful insights to future research in the field of workflow-guided agent planning.

\section*{Limitations}
We present FlowBench, a challenging benchmark for workflow-guided agent planning. Despite its comprehensiveness, this work has certain limitations that must be acknowledged. First, 
our benchmark covers three representative knowledge formats from varying abstraction levels,
but may not cover all potential formats, which we intend to explore in future endeavors. Moreover, we adopt GPT-4-Turbo for automatic assessment in several metrics during evaluation, which increases the evaluation cost of using our FlowBench. Our extraction process of workflow knowledge also hinges on human efforts to ensure the quality, which is quite expensive and time-consuming. Future endeavors could explore the automation of workflow-related knowledge extraction.

\section*{Ethics Statement}
In this paper, we introduce FlowBench, a
highly challenging multi-domain benchmark for evaluation workflow-guided agent planning abilities. During our benchmark construction, our human annotators are instructed to filter out data potentially triggering ethical concerns, including offensive content or social prejudices.
Even though the organized workflow knowledge and the ground-truth sessions have undergone filtering and verification, our simulated session-level evaluation still incorporates LLMs for simulating real-time interaction. Such API-based LLM simulators may exhibit bias and unfairness. We advise potential users to first apply bias reduction and correction methods to eliminate biased simulated sessions during evaluation, thereby enhancing fairness and ethical standards.
We will release our benchmark and evaluation scripts to foster innovation and aid the development of future research.

\bibliography{anthology,custom}

\begin{thebibliography}{42}
\providecommand{\natexlab}[1]{#1}

\bibitem[{Hel(2024)}]{HelloGPT4o}
 2024.
\newblock \href {https://openai.com/index/hello-gpt-4o/} {Hello {{GPT-4o}}}.

\bibitem[{Agrawal et~al.(2023)Agrawal, Kumarage, Alghami, and Liu}]{DBLP:journals/corr/abs-2311-07914}
Garima Agrawal, Tharindu Kumarage, Zeyad Alghami, and Huan Liu. 2023.
\newblock \href {https://doi.org/10.48550/ARXIV.2311.07914} {Can knowledge graphs reduce hallucinations in llms? : {A} survey}.
\newblock \emph{CoRR}, abs/2311.07914.

\bibitem[{Baek et~al.(2024)Baek, Chandrasekaran, Cucerzan, Herring, and Jauhar}]{DBLP:conf/www/BaekCCHJ24}
Jinheon Baek, Nirupama Chandrasekaran, Silviu Cucerzan, Allen Herring, and Sujay~Kumar Jauhar. 2024.
\newblock \href {https://doi.org/10.1145/3589334.3645404} {Knowledge-augmented large language models for personalized contextual query suggestion}.
\newblock In \emph{Proceedings of the {ACM} on Web Conference 2024, {WWW} 2024, Singapore, May 13-17, 2024}, pages 3355--3366. {ACM}.

\bibitem[{Chen et~al.(2023)Chen, Dong, Shu, Zhang, Sesay, Karlsson, Fu, and Shi}]{DBLP:journals/corr/abs-2309-17288}
Guangyao Chen, Siwei Dong, Yu~Shu, Ge~Zhang, Jaward Sesay, B{\"{o}}rje~F. Karlsson, Jie Fu, and Yemin Shi. 2023.
\newblock \href {https://doi.org/10.48550/ARXIV.2309.17288} {Autoagents: {A} framework for automatic agent generation}.
\newblock \emph{CoRR}, abs/2309.17288.

\bibitem[{Chen et~al.(2024)Chen, Du, Zhang, Liu, Liu, Zheng, Zhuo, Zhang, Lin, Chen, and Zhao}]{chen2024teval}
Zehui Chen, Weihua Du, Wenwei Zhang, Kuikun Liu, Jiangning Liu, Miao Zheng, Jingming Zhuo, Songyang Zhang, Dahua Lin, Kai Chen, and Feng Zhao. 2024.
\newblock \href {https://arxiv.org/abs/2312.14033} {T-eval: Evaluating the tool utilization capability of large language models step by step}.
\newblock \emph{Preprint}, arXiv:2312.14033.

\bibitem[{Chhikara et~al.(2023)Chhikara, Zhang, Ilievski, Francis, and Ma}]{DBLP:conf/kcap/ChhikaraZIFM23}
Prateek Chhikara, Jiarui Zhang, Filip Ilievski, Jonathan Francis, and Kaixin Ma. 2023.
\newblock \href {https://doi.org/10.1145/3587259.3627561} {Knowledge-enhanced agents for interactive text games}.
\newblock In \emph{Proceedings of the 12th Knowledge Capture Conference 2023, {K-CAP} 2023, Pensacola, FL, USA, December 5-7, 2023}, pages 157--165. {ACM}.

\bibitem[{Deng et~al.(2023)Deng, Gu, Zheng, Chen, Stevens, Wang, Sun, and Su}]{DBLP:conf/nips/DengGZCSWSS23}
Xiang Deng, Yu~Gu, Boyuan Zheng, Shijie Chen, Samual Stevens, Boshi Wang, Huan Sun, and Yu~Su. 2023.
\newblock \href {http://papers.nips.cc/paper\_files/paper/2023/hash/5950bf290a1570ea401bf98882128160-Abstract-Datasets\_and\_Benchmarks.html} {Mind2web: Towards a generalist agent for the web}.
\newblock In \emph{NeurIPS}.

\bibitem[{Gao et~al.(2023)Gao, Xiong, Gao, Jia, Pan, Bi, Dai, Sun, Guo, Wang, and Wang}]{DBLP:journals/corr/abs-2312-10997}
Yunfan Gao, Yun Xiong, Xinyu Gao, Kangxiang Jia, Jinliu Pan, Yuxi Bi, Yi~Dai, Jiawei Sun, Qianyu Guo, Meng Wang, and Haofen Wang. 2023.
\newblock \href {https://doi.org/10.48550/ARXIV.2312.10997} {Retrieval-augmented generation for large language models: {A} survey}.
\newblock \emph{CoRR}, abs/2312.10997.

\bibitem[{Hong et~al.(2023)Hong, Zheng, Chen, Cheng, Wang, Zhang, Wang, Yau, Lin, Zhou, Ran, Xiao, and Wu}]{DBLP:journals/corr/abs-2308-00352}
Sirui Hong, Xiawu Zheng, Jonathan Chen, Yuheng Cheng, Jinlin Wang, Ceyao Zhang, Zili Wang, Steven Ka~Shing Yau, Zijuan Lin, Liyang Zhou, Chenyu Ran, Lingfeng Xiao, and Chenglin Wu. 2023.
\newblock \href {https://doi.org/10.48550/ARXIV.2308.00352} {Metagpt: Meta programming for multi-agent collaborative framework}.
\newblock \emph{CoRR}, abs/2308.00352.

\bibitem[{Huang et~al.(2022)Huang, Abbeel, Pathak, and Mordatch}]{DBLP:conf/icml/HuangAPM22}
Wenlong Huang, Pieter Abbeel, Deepak Pathak, and Igor Mordatch. 2022.
\newblock \href {https://proceedings.mlr.press/v162/huang22a.html} {Language models as zero-shot planners: Extracting actionable knowledge for embodied agents}.
\newblock In \emph{International Conference on Machine Learning, {ICML} 2022, 17-23 July 2022, Baltimore, Maryland, {USA}}, volume 162 of \emph{Proceedings of Machine Learning Research}, pages 9118--9147. {PMLR}.

\bibitem[{Jiang et~al.(2024)Jiang, Zhou, Zhao, Song, Zhu, Zhu, and Wen}]{DBLP:journals/corr/abs-2402-11163}
Jinhao Jiang, Kun Zhou, Wayne~Xin Zhao, Yang Song, Chen Zhu, Hengshu Zhu, and Ji{-}Rong Wen. 2024.
\newblock \href {https://doi.org/10.48550/ARXIV.2402.11163} {Kg-agent: An efficient autonomous agent framework for complex reasoning over knowledge graph}.
\newblock \emph{CoRR}, abs/2402.11163.

\bibitem[{Koupaee and Wang(2018)}]{DBLP:journals/corr/abs-1810-09305}
Mahnaz Koupaee and William~Yang Wang. 2018.
\newblock \href {https://arxiv.org/abs/1810.09305} {Wikihow: {A} large scale text summarization dataset}.
\newblock \emph{CoRR}, abs/1810.09305.

\bibitem[{Liu et~al.(2023{\natexlab{a}})Liu, Zhou, Hua, Chong, Tian, Liu, Wang, You, Guo, Zhu, and Li}]{DBLP:conf/nips/LiuZHCTLWYGZL23}
Junling Liu, Peilin Zhou, Yining Hua, Dading Chong, Zhongyu Tian, Andrew Liu, Helin Wang, Chenyu You, Zhenhua Guo, Lei Zhu, and Michael~Lingzhi Li. 2023{\natexlab{a}}.
\newblock \href {http://papers.nips.cc/paper\_files/paper/2023/hash/a48ad12d588c597f4725a8b84af647b5-Abstract-Datasets\_and\_Benchmarks.html} {Benchmarking large language models on cmexam - {A} comprehensive chinese medical exam dataset}.
\newblock In \emph{NeurIPS}.

\bibitem[{Liu et~al.(2023{\natexlab{b}})Liu, Yu, Zhang, Xu, Lei, Lai, Gu, Ding, Men, Yang, Zhang, Deng, Zeng, Du, Zhang, Shen, Zhang, Su, Sun, Huang, Dong, and Tang}]{DBLP:journals/corr/abs-2308-03688}
Xiao Liu, Hao Yu, Hanchen Zhang, Yifan Xu, Xuanyu Lei, Hanyu Lai, Yu~Gu, Hangliang Ding, Kaiwen Men, Kejuan Yang, Shudan Zhang, Xiang Deng, Aohan Zeng, Zhengxiao Du, Chenhui Zhang, Sheng Shen, Tianjun Zhang, Yu~Su, Huan Sun, Minlie Huang, Yuxiao Dong, and Jie Tang. 2023{\natexlab{b}}.
\newblock \href {https://doi.org/10.48550/ARXIV.2308.03688} {Agentbench: Evaluating llms as agents}.
\newblock \emph{CoRR}, abs/2308.03688.

\bibitem[{Liu et~al.(2024)Liu, Wu, Xu, Chen, Zhang, Wang, and Gao}]{DBLP:journals/corr/abs-2404-00942}
Xiaoze Liu, Feijie Wu, Tianyang Xu, Zhuo Chen, Yichi Zhang, Xiaoqian Wang, and Jing Gao. 2024.
\newblock \href {https://doi.org/10.48550/ARXIV.2404.00942} {Evaluating the factuality of large language models using large-scale knowledge graphs}.
\newblock \emph{CoRR}, abs/2404.00942.

\bibitem[{Ma et~al.(2024)Ma, Zhang, Zhu, Yang, Yang, Jin, Lan, Kong, and He}]{DBLP:journals/corr/abs-2401-13178}
Chang Ma, Junlei Zhang, Zhihao Zhu, Cheng Yang, Yujiu Yang, Yaohui Jin, Zhenzhong Lan, Lingpeng Kong, and Junxian He. 2024.
\newblock \href {https://doi.org/10.48550/ARXIV.2401.13178} {Agentboard: An analytical evaluation board of multi-turn {LLM} agents}.
\newblock \emph{CoRR}, abs/2401.13178.

\bibitem[{Mosig et~al.(2020)Mosig, Mehri, and Kober}]{DBLP:journals/corr/abs-2010-11853}
Johannes E.~M. Mosig, Shikib Mehri, and Thomas Kober. 2020.
\newblock \href {https://arxiv.org/abs/2010.11853} {{STAR:} {A} schema-guided dialog dataset for transfer learning}.
\newblock \emph{CoRR}, abs/2010.11853.

\bibitem[{OpenAI(2022)}]{openai2022chatgpt}
OpenAI. 2022.
\newblock \href {https://openai.com/blog/chatgpt} {Chatgpt}.

\bibitem[{OpenAI(2023{\natexlab{a}})}]{DBLP:journals/corr/abs-2303-08774}
OpenAI. 2023{\natexlab{a}}.
\newblock \href {https://doi.org/10.48550/ARXIV.2303.08774} {{GPT-4} technical report}.
\newblock \emph{CoRR}, abs/2303.08774.

\bibitem[{OpenAI(2023{\natexlab{b}})}]{openai2023gpt4}
OpenAI. 2023{\natexlab{b}}.
\newblock Gpt-4 technical report.
\newblock \emph{arXiv preprint arXiv:2303.08774}.

\bibitem[{Patil et~al.(2023)Patil, Zhang, Wang, and Gonzalez}]{DBLP:journals/corr/abs-2305-15334}
Shishir~G. Patil, Tianjun Zhang, Xin Wang, and Joseph~E. Gonzalez. 2023.
\newblock \href {https://doi.org/10.48550/ARXIV.2305.15334} {Gorilla: Large language model connected with massive apis}.
\newblock \emph{CoRR}, abs/2305.15334.

\bibitem[{Qian et~al.(2023)Qian, Cong, Yang, Chen, Su, Xu, Liu, and Sun}]{DBLP:journals/corr/abs-2307-07924}
Chen Qian, Xin Cong, Cheng Yang, Weize Chen, Yusheng Su, Juyuan Xu, Zhiyuan Liu, and Maosong Sun. 2023.
\newblock \href {https://doi.org/10.48550/ARXIV.2307.07924} {Communicative agents for software development}.
\newblock \emph{CoRR}, abs/2307.07924.

\bibitem[{Qin et~al.(2023)Qin, Liang, Ye, Zhu, Yan, Lu, Lin, Cong, Tang, Qian, Zhao, Tian, Xie, Zhou, Gerstein, Li, Liu, and Sun}]{DBLP:journals/corr/abs-2307-16789}
Yujia Qin, Shihao Liang, Yining Ye, Kunlun Zhu, Lan Yan, Yaxi Lu, Yankai Lin, Xin Cong, Xiangru Tang, Bill Qian, Sihan Zhao, Runchu Tian, Ruobing Xie, Jie Zhou, Mark Gerstein, Dahai Li, Zhiyuan Liu, and Maosong Sun. 2023.
\newblock \href {https://doi.org/10.48550/ARXIV.2307.16789} {Toolllm: Facilitating large language models to master 16000+ real-world apis}.
\newblock \emph{CoRR}, abs/2307.16789.

\bibitem[{Schick et~al.(2023)Schick, Dwivedi{-}Yu, Dess{\`{\i}}, Raileanu, Lomeli, Hambro, Zettlemoyer, Cancedda, and Scialom}]{DBLP:conf/nips/SchickDDRLHZCS23}
Timo Schick, Jane Dwivedi{-}Yu, Roberto Dess{\`{\i}}, Roberta Raileanu, Maria Lomeli, Eric Hambro, Luke Zettlemoyer, Nicola Cancedda, and Thomas Scialom. 2023.
\newblock \href {http://papers.nips.cc/paper\_files/paper/2023/hash/d842425e4bf79ba039352da0f658a906-Abstract-Conference.html} {Toolformer: Language models can teach themselves to use tools}.
\newblock In \emph{NeurIPS}.

\bibitem[{Shi et~al.(2023)Shi, Min, Yasunaga, Seo, James, Lewis, Zettlemoyer, and Yih}]{DBLP:journals/corr/abs-2301-12652}
Weijia Shi, Sewon Min, Michihiro Yasunaga, Minjoon Seo, Rich James, Mike Lewis, Luke Zettlemoyer, and Wen{-}tau Yih. 2023.
\newblock \href {https://doi.org/10.48550/ARXIV.2301.12652} {{REPLUG:} retrieval-augmented black-box language models}.
\newblock \emph{CoRR}, abs/2301.12652.

\bibitem[{Touvron et~al.(2023)Touvron, Lavril, Izacard, Martinet, Lachaux, Lacroix, Rozi{\`{e}}re, Goyal, Hambro, Azhar, Rodriguez, Joulin, Grave, and Lample}]{DBLP:journals/corr/abs-2302-13971}
Hugo Touvron, Thibaut Lavril, Gautier Izacard, Xavier Martinet, Marie{-}Anne Lachaux, Timoth{\'{e}}e Lacroix, Baptiste Rozi{\`{e}}re, Naman Goyal, Eric Hambro, Faisal Azhar, Aur{\'{e}}lien Rodriguez, Armand Joulin, Edouard Grave, and Guillaume Lample. 2023.
\newblock \href {https://doi.org/10.48550/ARXIV.2302.13971} {Llama: Open and efficient foundation language models}.
\newblock \emph{CoRR}, abs/2302.13971.

\bibitem[{Valmeekam et~al.(2023)Valmeekam, Marquez, Hernandez, Sreedharan, and Kambhampati}]{DBLP:conf/nips/ValmeekamMHSK23}
Karthik Valmeekam, Matthew Marquez, Alberto~Olmo Hernandez, Sarath Sreedharan, and Subbarao Kambhampati. 2023.
\newblock \href {http://papers.nips.cc/paper\_files/paper/2023/hash/7a92bcdede88c7afd108072faf5485c8-Abstract-Datasets\_and\_Benchmarks.html} {Planbench: An extensible benchmark for evaluating large language models on planning and reasoning about change}.
\newblock In \emph{NeurIPS}.

\bibitem[{Wang et~al.(2024)Wang, Ma, Feng, Zhang, Yang, Zhang, Chen, Tang, Chen, Lin, Zhao, Wei, and Wen}]{DBLP:journals/fcsc/WangMFZYZCTCLZWW24}
Lei Wang, Chen Ma, Xueyang Feng, Zeyu Zhang, Hao Yang, Jingsen Zhang, Zhiyuan Chen, Jiakai Tang, Xu~Chen, Yankai Lin, Wayne~Xin Zhao, Zhewei Wei, and Jirong Wen. 2024.
\newblock \href {https://doi.org/10.1007/S11704-024-40231-1} {A survey on large language model based autonomous agents}.
\newblock \emph{Frontiers Comput. Sci.}, 18(6):186345.

\bibitem[{Wei et~al.(2022)Wei, Wang, Schuurmans, Bosma, Ichter, Xia, Chi, Le, and Zhou}]{DBLP:conf/nips/Wei0SBIXCLZ22}
Jason Wei, Xuezhi Wang, Dale Schuurmans, Maarten Bosma, Brian Ichter, Fei Xia, Ed~H. Chi, Quoc~V. Le, and Denny Zhou. 2022.
\newblock \href {http://papers.nips.cc/paper\_files/paper/2022/hash/9d5609613524ecf4f15af0f7b31abca4-Abstract-Conference.html} {Chain-of-thought prompting elicits reasoning in large language models}.
\newblock In \emph{NeurIPS}.

\bibitem[{Wu et~al.(2023)Wu, Tang, Mitchell, and Li}]{DBLP:journals/corr/abs-2310-01557}
Yue Wu, Xuan Tang, Tom~M. Mitchell, and Yuanzhi Li. 2023.
\newblock \href {https://doi.org/10.48550/ARXIV.2310.01557} {Smartplay : {A} benchmark for llms as intelligent agents}.
\newblock \emph{CoRR}, abs/2310.01557.

\bibitem[{Xie et~al.(2024)Xie, Zhang, Chen, Zhu, Lou, Tian, Xiao, and Su}]{DBLP:journals/corr/abs-2402-01622}
Jian Xie, Kai Zhang, Jiangjie Chen, Tinghui Zhu, Renze Lou, Yuandong Tian, Yanghua Xiao, and Yu~Su. 2024.
\newblock \href {https://doi.org/10.48550/ARXIV.2402.01622} {Travelplanner: {A} benchmark for real-world planning with language agents}.
\newblock \emph{CoRR}, abs/2402.01622.

\bibitem[{Xu et~al.(2024)Xu, He, Chen, Wang, Song, Tong, Liu, and Zhao}]{DBLP:journals/corr/abs-2404-14741}
Yao Xu, Shizhu He, Jiabei Chen, Zihao Wang, Yangqiu Song, Hanghang Tong, Kang Liu, and Jun Zhao. 2024.
\newblock \href {https://doi.org/10.48550/ARXIV.2404.14741} {Generate-on-graph: Treat {LLM} as both agent and {KG} in incomplete knowledge graph question answering}.
\newblock \emph{CoRR}, abs/2404.14741.

\bibitem[{Yang et~al.(2024{\natexlab{a}})Yang, Liu, Wu, Yang, Fung, Li, Huang, Cao, Wang, Wang, Ji, and Zhai}]{DBLP:journals/corr/abs-2401-00812}
Ke~Yang, Jiateng Liu, John Wu, Chaoqi Yang, Yi~R. Fung, Sha Li, Zixuan Huang, Xu~Cao, Xingyao Wang, Yiquan Wang, Heng Ji, and Chengxiang Zhai. 2024{\natexlab{a}}.
\newblock \href {https://doi.org/10.48550/ARXIV.2401.00812} {If {LLM} is the wizard, then code is the wand: {A} survey on how code empowers large language models to serve as intelligent agents}.
\newblock \emph{CoRR}, abs/2401.00812.

\bibitem[{Yang et~al.(2024{\natexlab{b}})Yang, Zhao, Zhu, Zhou, Xu, Jia, and Zan}]{DBLP:conf/aaai/YangZZZXJZ24}
Songhua Yang, Hanjie Zhao, Senbin Zhu, Guangyu Zhou, Hongfei Xu, Yuxiang Jia, and Hongying Zan. 2024{\natexlab{b}}.
\newblock \href {https://doi.org/10.1609/AAAI.V38I17.29907} {Zhongjing: Enhancing the chinese medical capabilities of large language model through expert feedback and real-world multi-turn dialogue}.
\newblock In \emph{AAAI}, pages 19368--19376. {AAAI} Press.

\bibitem[{Yao et~al.(2023{\natexlab{a}})Yao, Yu, Zhao, Shafran, Griffiths, Cao, and Narasimhan}]{DBLP:conf/nips/YaoYZS00N23}
Shunyu Yao, Dian Yu, Jeffrey Zhao, Izhak Shafran, Tom Griffiths, Yuan Cao, and Karthik Narasimhan. 2023{\natexlab{a}}.
\newblock \href {http://papers.nips.cc/paper\_files/paper/2023/hash/271db9922b8d1f4dd7aaef84ed5ac703-Abstract-Conference.html} {Tree of thoughts: Deliberate problem solving with large language models}.
\newblock In \emph{NeurIPS}.

\bibitem[{Yao et~al.(2023{\natexlab{b}})Yao, Zhao, Yu, Du, Shafran, Narasimhan, and Cao}]{DBLP:conf/iclr/YaoZYDSN023}
Shunyu Yao, Jeffrey Zhao, Dian Yu, Nan Du, Izhak Shafran, Karthik~R. Narasimhan, and Yuan Cao. 2023{\natexlab{b}}.
\newblock \href {https://openreview.net/pdf?id=WE\_vluYUL-X} {React: Synergizing reasoning and acting in language models}.
\newblock In \emph{ICLR}. OpenReview.net.

\bibitem[{Ye et~al.(2023)Ye, Cong, Tian, Cao, Wang, Qin, Lu, Yu, Wang, Lin, Liu, and Sun}]{DBLP:journals/corr/abs-2311-10751}
Yining Ye, Xin Cong, Shizuo Tian, Jiannan Cao, Hao Wang, Yujia Qin, Yaxi Lu, Heyang Yu, Huadong Wang, Yankai Lin, Zhiyuan Liu, and Maosong Sun. 2023.
\newblock \href {https://doi.org/10.48550/ARXIV.2311.10751} {Proagent: From robotic process automation to agentic process automation}.
\newblock \emph{CoRR}, abs/2311.10751.

\bibitem[{Zapier()}]{zapier}
Zapier.
\newblock \href {https://zapier.com/} {Zapier | automation makes you move forward.}

\bibitem[{Zeng et~al.(2023)Zeng, Liu, Du, Wang, Lai, Ding, Yang, Xu, Zheng, Xia, Tam, Ma, Xue, Zhai, Chen, Liu, Zhang, Dong, and Tang}]{DBLP:conf/iclr/ZengLDWL0YXZXTM23}
Aohan Zeng, Xiao Liu, Zhengxiao Du, Zihan Wang, Hanyu Lai, Ming Ding, Zhuoyi Yang, Yifan Xu, Wendi Zheng, Xiao Xia, Weng~Lam Tam, Zixuan Ma, Yufei Xue, Jidong Zhai, Wenguang Chen, Zhiyuan Liu, Peng Zhang, Yuxiao Dong, and Jie Tang. 2023.
\newblock \href {https://openreview.net/pdf?id=-Aw0rrrPUF} {{GLM-130B:} an open bilingual pre-trained model}.
\newblock In \emph{ICLR}. OpenReview.net.

\bibitem[{Zhang et~al.(2023)Zhang, Chen, Jiang, Yu, Chen, Chen, Li, Wu, Zhiyi, Xiao, Wan, Wang, and Li}]{zhang-etal-2023-huatuogpt}
Hongbo Zhang, Junying Chen, Feng Jiang, Fei Yu, Zhihong Chen, Guiming Chen, Jianquan Li, Xiangbo Wu, Zhang Zhiyi, Qingying Xiao, Xiang Wan, Benyou Wang, and Haizhou Li. 2023.
\newblock \href {https://doi.org/10.18653/v1/2023.findings-emnlp.725} {{H}uatuo{GPT}, towards taming language model to be a doctor}.
\newblock In \emph{Findings of the Association for Computational Linguistics: EMNLP 2023}, pages 10859--10885, Singapore. Association for Computational Linguistics.

\bibitem[{Zhou et~al.(2023)Zhou, Xu, Zhu, Zhou, Lo, Sridhar, Cheng, Bisk, Fried, Alon, and Neubig}]{DBLP:journals/corr/abs-2307-13854}
Shuyan Zhou, Frank~F. Xu, Hao Zhu, Xuhui Zhou, Robert Lo, Abishek Sridhar, Xianyi Cheng, Yonatan Bisk, Daniel Fried, Uri Alon, and Graham Neubig. 2023.
\newblock \href {https://doi.org/10.48550/ARXIV.2307.13854} {Webarena: {A} realistic web environment for building autonomous agents}.
\newblock \emph{CoRR}, abs/2307.13854.

\bibitem[{Zhu et~al.(2024)Zhu, Qiao, Ou, Deng, Zhang, Lyu, Shen, Liang, Gu, and Chen}]{DBLP:journals/corr/abs-2403-03101}
Yuqi Zhu, Shuofei Qiao, Yixin Ou, Shumin Deng, Ningyu Zhang, Shiwei Lyu, Yue Shen, Lei Liang, Jinjie Gu, and Huajun Chen. 2024.
\newblock \href {https://doi.org/10.48550/ARXIV.2403.03101} {Knowagent: Knowledge-augmented planning for llm-based agents}.
\newblock \emph{CoRR}, abs/2403.03101.

\end{thebibliography}

\appendix

\begin{table*}[!t]
    \centering
    \small
    \renewcommand\arraystretch{1.1}
    \tabcolsep=0.12cm
    \begin{tabular}{l|l|l}
        \toprule
        \textbf{Domain} & \textbf{Roles} & \textbf{Scenarios}
      \\
        \midrule
    \multirow{4}{*}{\textbf{Customer Service}} &  {restaurant\_waiter}  & {Restaurant Search, Restaurant Booking}   \\
    & {hotel\_reception}  & {Hotel Search, Hotel Booking, Hotel Room Service} \\
    & {apartment\_manager}  & {Apartment Search, Schedule a Viewing, Bill Inquiry, Rent Payment} \\
    & {gas\_equipment\_service}  & {Gas Repairs, Gas Bill Payment, Gas Interruption Feedback} \\ \midrule
    
    \multirow{3}{*}{\textbf{Personal Assistant}} & {medical\_consultant}  & {Schedule a Medical Consultation, Obtain Diagnostic Results}  \\
    & meeting\_arrangement & {Meeting Initiation, Meeting Reschedule} \\
    & financial\_assistant & {Currency Exchange, Withdrawal Appointments, Balance Inquiry}\\ \midrule
    
    \multirow{2}{*}{\textbf{E-tailing Recommandation}} &   {online\_shopping\_support}  & {Product Search, Cart Management, Order Processing} \\
&   {computer\_store\_sale}  & {Laptop Recommendations, Computer Maintenance} \\ \midrule

    \multirow{4}{*}{\textbf{Travel\&Transportation}} &   
    {ride\_service}  & {Ride Booking, Ride Inquiry} \\
& {driving\_service}  & {Driving Consultation, Cancel Reservation, Modify Reservation} \\
& {flight\_inquiry}  & {Flight Information Search, Flight Booking} \\
& {travel\_assistant}  & {Travel Guidance, Weather Inquiry} \\ \midrule

    \multirow{3}{*}{\textbf{Logistics Solutions}} &  
    {express\_support}  & {Express Delivery, Express Tracking Inquiry} \\
    & {moving\_service}  & {Moving Service Appointment, Insurance Claim} \\
    & {food\_delivery\_service}  & {Online Questionnaires, Get Food Voucher} \\ \midrule
    
    \multirow{6}{*}{\textbf{Robotic Process Automation}} &  
     {invoice\_management}  & {Invoice Administration, Invoice Reimbursement} \\
    & {mail\_administration}  & {Mail transmission, Mail Response} \\  
    &{printing\_service}  & {Document Printing, Printing State Notification} \\
    & {attendance\_arrangement}  & {Attendance Anomaly Detection, Shift Handover} \\
    & {seal\_management}  & {Seal Request, Seal State Notification} \\
    & {workstation\_applicant}  & {Workstation Replacement, Workstation Change Directive} \\

    \bottomrule
       
    \end{tabular}
    \caption{Detailed Scenarios of different domains and roles for FlowBench. }
    \label{tab:app_scenario_details}
\end{table*}
\section{Additional Benchmark Details}

\subsection{Entire List of Roles and Scenarios}
As mentioned in Section~\ref{sec:bench_con}, we 
collect our tasks from six separate domains, with each domain encompassing different roles and scenarios. Here we provide the entire of roles and scenarios within each domain, as shown in Table~\ref{tab:app_scenario_details}.

\subsection{Entire Benchmark Statistics}
Here we further provide the entire benchmark statistics in Table~\ref{tab:app_domain_sta}, including the number of sessions and turns under the different settings of single-scenario and cross-scenario respectively.

\subsection{Benchmark Construction Cost}\label{app:cons_cost}
We further provide the rough cost of benchmark construction. Before quality filtering, we manually constructed a total of 58 scenarios and 730 dialogues. The annotation cost is \$17.23 per scenario and \$9.44 per dialogue, amounting to a total annotation cost of \$7,892.60. Additionally, the cost for generating multi-turn dialogue requests using GPT-4-Turbo is approximately \$1,000, bringing the total cost to around \$8,892.60.

\subsection{Formatting Standards}
In the process of organizing workflow knowledge, we need to convert the text-based natural language documents of workflow knowledge into code and flowchart formats. Different standard requirements are maintained for them. (i)-For code format, we adopt Python-style pseudo code, mandating that various action steps be delineated within distinct functions. The variable names and parameters must correspond to the workflow implications from text documents. For commands that are difficult to express through code, minimal commenting is permitted for clarification. (ii)-For flowchart format, we utilize Markdown Mermaid syntax for its visual operation capabilities. In the flowcharts, nodes correspond to distinct states, outlining within them the workflow directives applicable in those states, which entails rules of actions and responses. Edges represent the possible movements between nodes, with transition conditions based on user intents or environment feedback.

\subsection{Demonstration of Samples}
We provide a demonstration of samples, including different knowledge formats, tool information and complete ground-truth sessions in Appendix~\ref{app:add_demo}.

\section{Additional Experimental Results}\label{app:results}

\begin{figure}[!t]
     \centering
     \begin{subfigure}{0.49\linewidth}
         \includegraphics[width=\columnwidth]{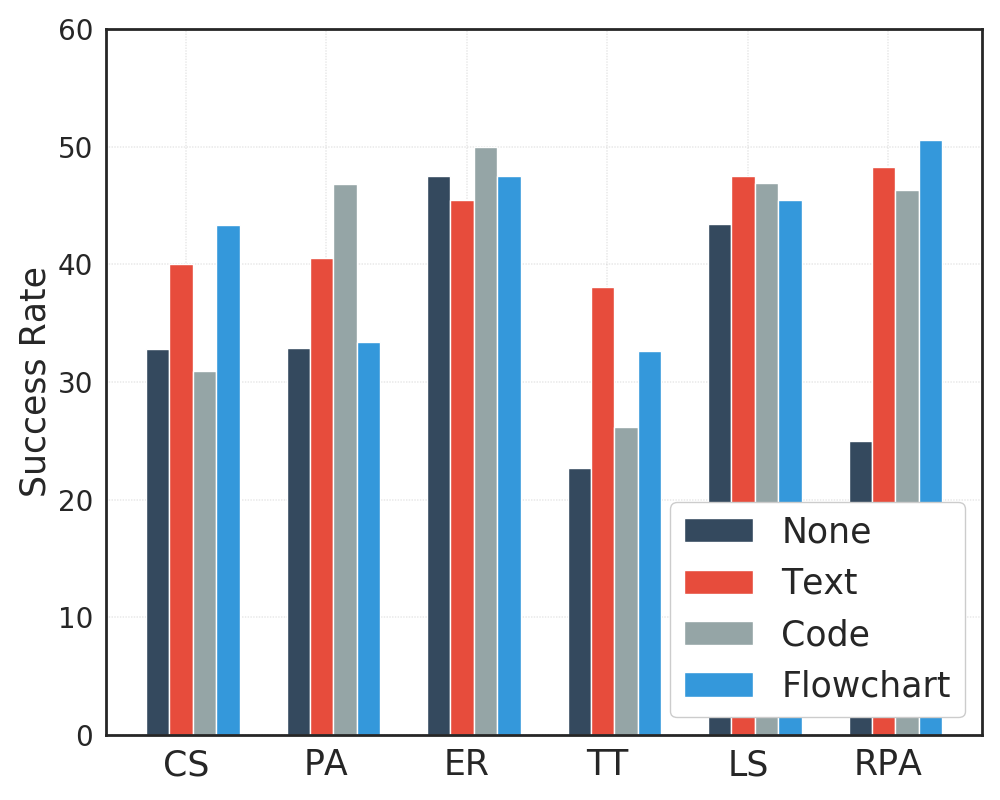}
         \caption{Single-Scenario.}
     \end{subfigure}
     \hfill
     \begin{subfigure}{0.49\linewidth}
         \includegraphics[width=\columnwidth]{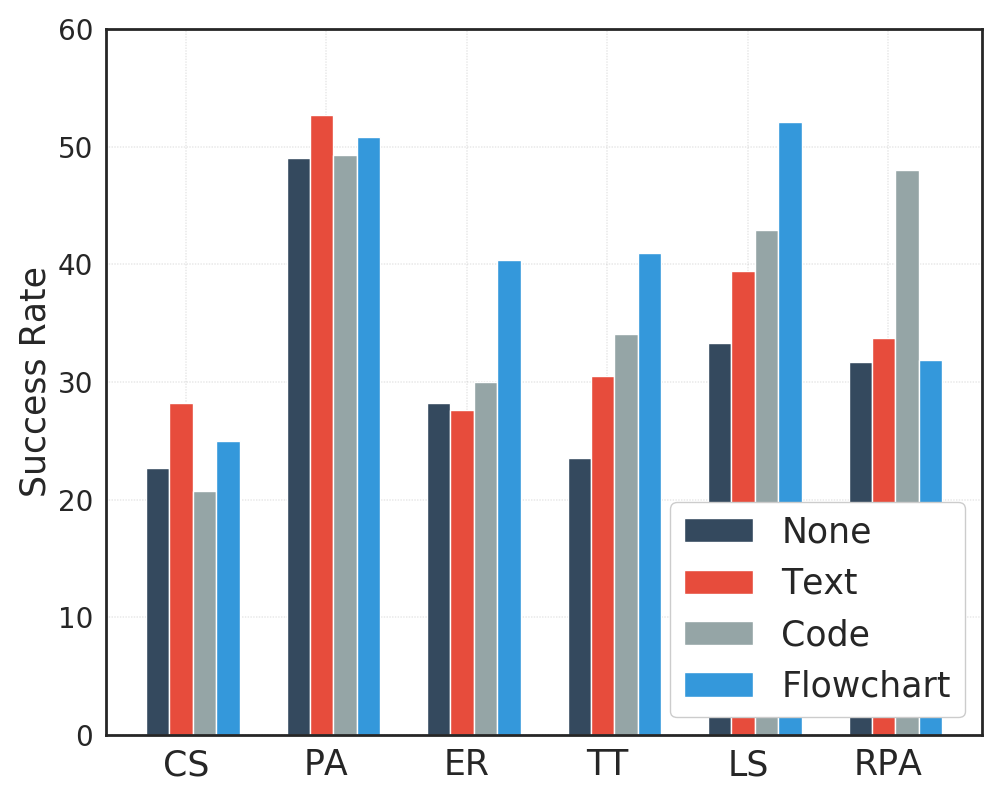}
         \caption{Cross-Scenario. }
     \end{subfigure}
     
     \caption{Domain-wise success rate under session-level evaluation. Initials are adopted to denote domains. }
     \label{fig:app_distri_sr}
\end{figure}

\subsection{Additional Performance Distribution}
We further additional performance distribution across domains. For experiments on GPT-4-Turbo, the domain-wise success rate is depicted in Figure~\ref{fig:app_distri_sr}. It can be observed that workflow knowledge has a substantial impact in some expertise-demanding domains such as RPA, E-tailing Recommendation, and Travel\&Tranportation. Some domains show a noticeable performance drop in the cross-scenario setting compared to the single-scenario setting. We speculate this is due to a larger scenario gap within the domain.

\begin{table}[t]

    \renewcommand{\arraystretch}{1.0}
    \tabcolsep=0.17cm
    \small
    \centering
    \begin{tabular}{l|l|ccccc}
    \toprule
    \multirow{2}{*}{\begin{tabular}[c]{@{}c@{}}Scenario \\ Setting\end{tabular}} & \multirow{2}{*}{Format}& \multicolumn{5}{c}{Type of Failure Reasons} 
    \\
    \addlinespace[2pt]
    \cline{3-7}
    \addlinespace[3pt]
     & & {1} & {2} & {3} 
     & {4}  & {5}
    \\ %
    \midrule
\multirow{4}{*}{\textbf{Single}} & None & 52.1 & 4.5 & 6.0 & 35.1 & 2.3 \\ 
& {Text} &  52.2 & 3.3 & 2.2 & 40.2 & 2.1   \\
& {Code} &  57.3 & 1.9 & 2.9 & 35.9 & 2.0 \\
&{Flowchart} & 56.9 & 0.9 & 3.9 & 34.9 & 3.4 \\
\midrule
\multirow{4}{*}{\textbf{Cross}} & None & 32.4 & 4.3 & 9.0 & 54.3 & 0.0 \\ 
& {Text} &  45.8 & 1.3 & 4.2 & 46.7 & 2.0   \\
& {Code} &  41.4 & 1.2 & 8.0 & 47.1 & 2.3 \\
&{Flowchart} & 49.5 & 0.9 & 7.5 & 40.9 & 1.2 \\
\midrule

\end{tabular}
\caption{Analysis of different failure reasons for session-level evaluations with GPT-4-Turbo.
This shows the percentage share (\%) of different failure reasons.
}
\label{tab:app_causes}
\end{table}

\subsection{Analysis of Failure Causes}
We further analyze the reasons for task failure in session-level evaluation. Based on the workflow content and observations of the agent predictions, we categorize the failure reasons into the following five categories,
(i)-Type 1: Missing steps, which means the agent missed a step needed to complete the task.
(ii)-Type 2: Incorrect sequence, which means the agent employs incorrect order of multiple steps.
(iii)-Type 3: Incorrect transition, which includes incorrect recognition of user intent and transition conditions.
(iv)-Type 4: Tool usage, which means the agent fails in calling the correct tool name or in collecting parameters.
(v)-Type 5: Other reasons.
We conduct statistics on the reasons for failure in the samples via GPT-4-Turbo. Table~\ref{tab:app_causes} illustrates the proportion (\%) of failure causes in each setting for GPT-4-Turbo. It can be observed that lacking workflow guidance often leads to an increase in step sequencing (type 2) and transition errors (type 3), while flowchart formats are effective in addressing sequence errors. Further, with more tools available in cross-scenario situations, there's also an increase in tool invocation errors.

\begin{table*}[!t]
    \centering
    \small
    \renewcommand\arraystretch{1.1}
    \tabcolsep=0.12cm
    \begin{tabular}{l|cc|cc|cc}
        \toprule
       \multirow{2}{*}{\textbf{Domain}} & \multirow{2}{*}{\textbf{Roles}} & \multirow{2}{*}{\textbf{Scenarios}} 
      & \multicolumn{2}{|c}{\textbf{Sessions}} & \multicolumn{2}{|c}{\textbf{Turns}} \\
      &&& Single-Scenario & Cross-Scenario & Single-Scenario & Cross-Scenario \\
        \midrule
    {Customer Service} &   {4}  & {12}  & {66} & {48} & {619}& {547}\\
    {Personal Assistant} &   {3}  & {7}  & {59} & {33} & {510} & {311}\\
    {E-tailing Recommandation} &   {2}  & {5}  &  {21} & {11} & {199} & {131}\\
    {Travel\&Transportation} &   {4}  & {9}  &  {78} & {57} & {764} & {657} \\
    {Logistics Solutions} &   {3}  & {6}   & {38} & {23} & {299} & {222} \\
    {Robotic Process Automation} &   {6}  & {12}  & {73} & {29} & 703 & 350 \\
    \midrule
    {Total} & 22 & 51 & 335 & 201 & 3094 & 2219  \\
        \bottomrule
    \end{tabular}
    \caption{Entire domain statistics of FlowBench. }
    \label{tab:app_domain_sta}
\end{table*}

\begin{table}[!t]
    \centering
    \small
    \renewcommand\arraystretch{1.1}
    \tabcolsep=0.13cm
    \begin{tabular}{l|ccc}  
        \toprule
        \textbf{Task}    
      & {GPT-4o} & {GPT-4-Turbo} & {GPT-3.5-Turbo}  \\
        \midrule
        {Single-Scenario}  & {92.5} & {91.8} & {77.9} \\ 
        {Cross-Scenario}  & 84.0 & 70.7 & 57.3 
        \\
        \bottomrule
    \end{tabular}
    \caption{Comparisons of node prediction accuracy for turn-level flowchart-guided evaluation.}
    \label{tab:app_nodeacc}
\end{table}

\subsection{Average Number of Turns}
\begin{figure}[!t]
     \centering
     \begin{subfigure}{0.49\linewidth}
         \includegraphics[width=\columnwidth]{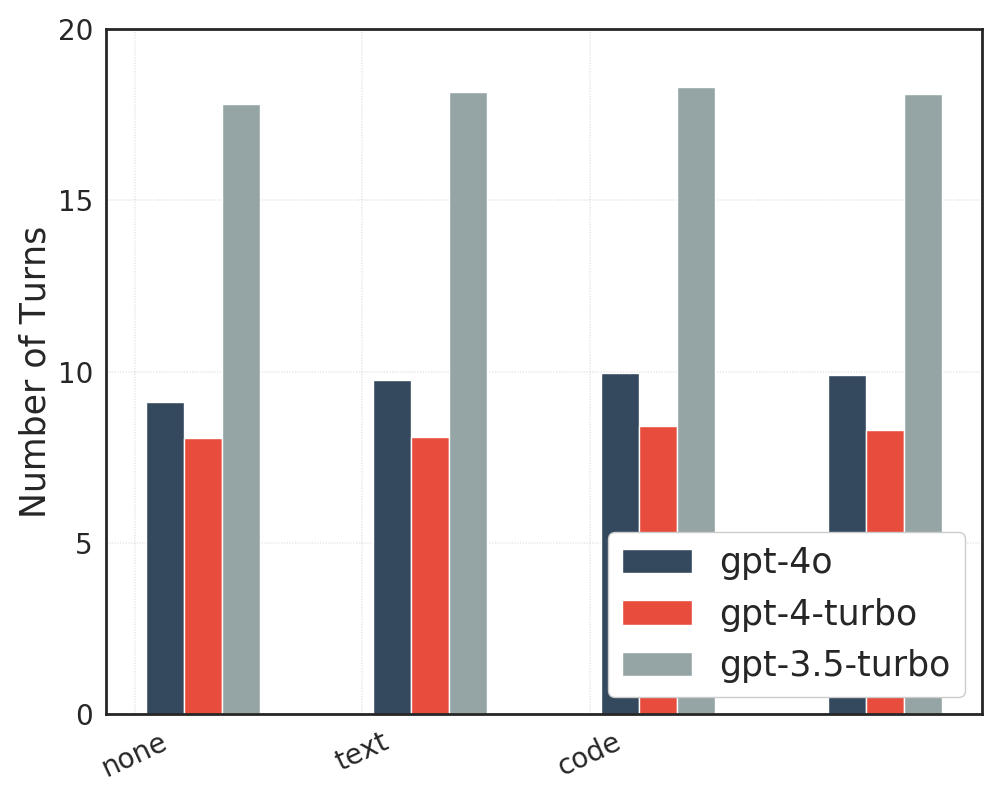}
         \caption{Single-Scenario.}
     \end{subfigure}
     \hfill
     \begin{subfigure}{0.49\linewidth}
         \includegraphics[width=\columnwidth]{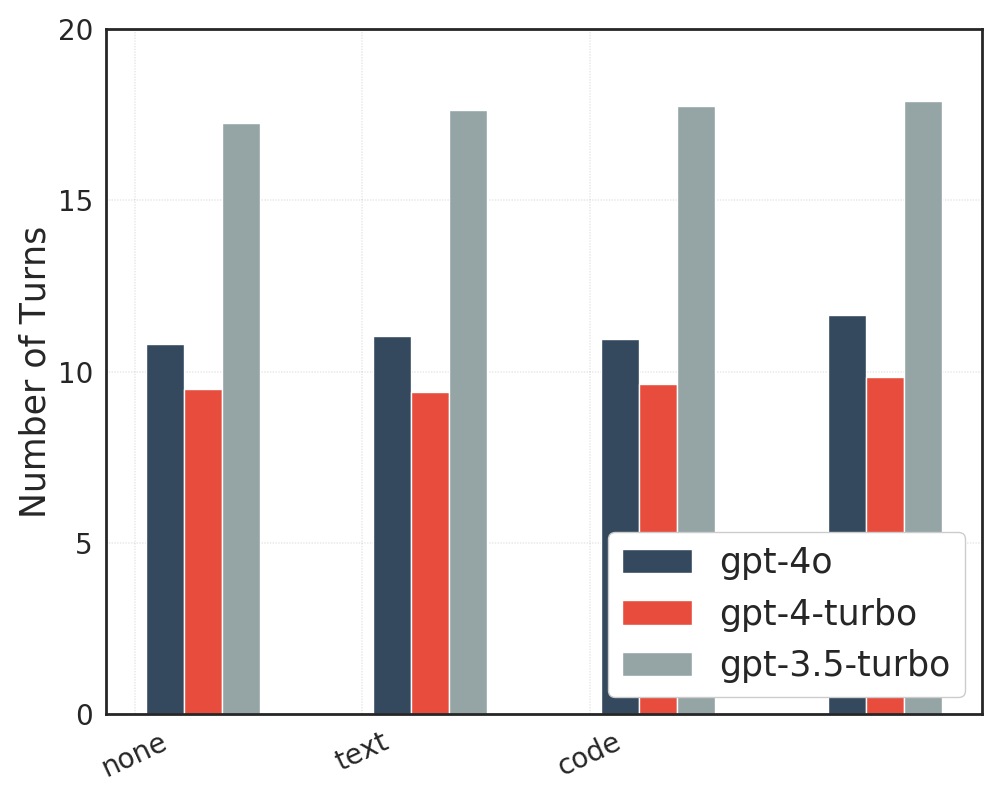}
         \caption{Cross-Scenario. }
     \end{subfigure}
     
     \caption{Average number of generated turns for simulated sessions under session-level evaluation. }
     \label{fig:app_turns}
\end{figure}

We further display the average number of turns for the simulated sessions during session-level evaluation in Figure~\ref{fig:app_turns} (each turn consisting of either an agent-user or agent-environment interaction). It can be observed that the lack of workflow results in slightly fewer rounds required (since the agent is utterly clueless about what actions to perform), whereas various formats have a negligible impact. For different LLMs, GPT-3.5-Turbo demands a larger amount of turns, with cross-scenario cases typically requiring more rounds than single-scenario as well.

\subsection{Results of Node Prediction}
Due to the special graph-structured nature of flowcharts, more analysis results are possible. For flowchart-guided planning, every step involves moving to a state node. We provide node accuracy metrics for turn-level flowchart-guided planning, illustrated in Table~\ref{tab:app_nodeacc}.  It can be observed that the most capable LLM GPT-4o can achieve satisfactory performance, while there is a significant variation among LLMs of different capabilities, indicating substantial differences in their ability to comprehend structured flowcharts.

\section{Additional Implementation Details}\label{app:more_imple}
\subsection{More Implementation Details}
In the construction process of our benchmark, 
we filtered out 12\% of the scenarios and 26\% of the sessions after the initial annotation.
During the ground-truth session generation, we also follow the setting of task awareness and generate sessions of both single-scenario and cross-scenario settings. In turn-level assessments, all ground-truth sessions are utilized. For simulated session-level evaluation, however, we preserve only those portions of the ground-truth sessions that are lengthy and more comprehensive, and then a user-task summary is created for simulation. For session-level assessment, the maximum number of interaction turns is restricted to 15 for single-scenario and 20 for cross-scenario. If the task is not completed within these limits, it is deemed a failure.

Moreover, we further provide additional evaluation details. During the evaluation of tool invocation, parameter matching is required. For types like numbers and enumerations, an exact match between predictions and ground truth is employed. In cases of more substantial strings where similar phrases could imply the same idea, we resort to using GPT-4-turbo for fuzzy matching, guided by precedent work. During our evaluations, All experiments are conducted three times, and the mean results are displayed. 

\begin{figure}[!t]
     \centering
         \includegraphics[width=0.95\columnwidth]{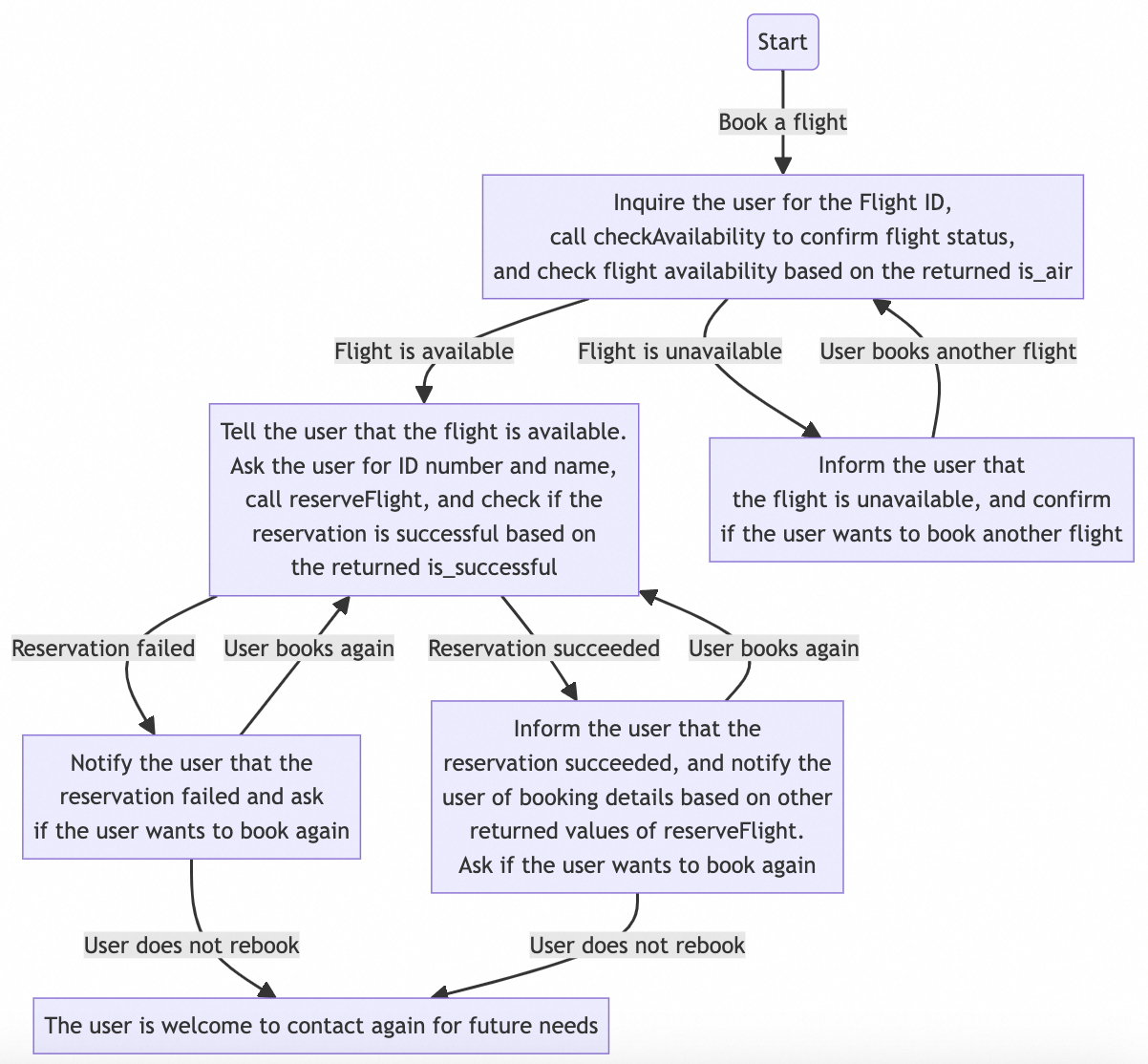}
     \caption{Visualization of flowchart format workflow knowledge for the scenario of flight booking.} 
     \label{fig:app_visflowchart}
\end{figure}

\subsection{Prompt Design for Inference}
Below we provide the inference prompt with the ReAct strategy. Notably, for both turn-level and session-level evaluations, the same inference prompt is used. For the cross-scenario setting, all the task-related knowledge bases from relevant scenarios are sequentially inserted into the prompt.

\begin{lstlisting}[language={}, caption=Prompt for inference]
You are a helpful assistant for the task of ..... 

{task background}

Specific requirements:
1. You need to act as an assistant and engage in a conversation with the user, following the business process and API information.
2. You have been provided with the flowchart information for different scenarios under a specific role. 
3. You can only answer questions within the scope of the given several workflow processes. If the user asks a question beyond these scopes, please apologize and explain to the user in the response part.
4. When asking for API input parameters, ensure that the provided parameter values comply with the specified format regarding both the correctness of the format and the completeness of the content. Do not assign values arbitrarily. In instances where the parameters do not meet the format requirements, notify users to make the necessary adjustments until the requirements are satisfied.
5. When the user has multiple requests at the same time, please select one appropriate request for processing first and inform the user that other requests will be resolved subsequently. If there is unfinished business in the previous conversation, continue to provide the necessary help and guidance to assist them in completing the business process. When multiple APIs need to be called, do so in separate rounds, with a maximum of one API call output per round. When the user indicates that the business is finished or says goodbye, respond politely and end the conversation.
6. Your output format should be chosen from one of the two templates below (7.1 and 7.2):
7.1 If you need to interact with the user:
```
Thought: xxx (description of your thought process )
Response: xxx (the content you need to inquire or reply)
```
[Format Explanation]
(1) Thought includes 2 pieces of information: [Step: Analyze the current intent]: 'The current intent intent=xxx'. [Step: Decide the follow-up actions]: 'Next, I need to xxx.'
7.2 If you need to call an API (only one API call per time):
```
Thought: xxx (description of your thought process )
Action: xxx (the function name to be called, do not add the prefix "functions.")
Action Input: xxx (the parameters for the function, must be in strict JSON format)
```
[Format Explanation]
(1) 'Thought' includes the information described in sections (1) and (3) of 7.1, totaling seven [Step: xxx] pieces of information.
(2) In template 7.2, do not output 'Response', only output 'Thought', 'Action', and 'Action Input'.

8. When multiple possible intents exist, connect them with ' OR ' and clarify which situation is being inquired about; when expressing multiple intents, connect them with ' AND '.

Please adhere strictly to the defined output format. Do not output the placeholder "...." verbatim; instead, fill it with the corresponding content.

Workflow information:
{workflow information in specified formats}

Tool information:
{toolbox information}

Current time:
{current time}

\end{lstlisting}

\subsection{Prompt Design for Automated Evaluation.}\label{app:prompt_eval}

\paragraph{Response Scoring.}
Below we provide the evaluation prompt of response scoring for turn-level evaluations. The full score is 10 points.
\begin{lstlisting}[language={}, caption=Prompt for response scoring]
Please serve as an impartial judge to evaluate the response quality of the assistant. Your evaluation should be based on the following criteria:
(1) Correctness: Does the reply remain consistent with the workflow knowledge without any contradictions?
(2) Helpfulness: Has the user's request been reasonably understood and addressed, fulfilling the user's needs within the provided workflow scope?
(3) Humanness: Is the response coherent, clear, complete, and does it include human acknowledgment?
Please compare the provided response with the reference response and evaluate it based on the mentioned dimensions. Then, aggregate these assessments to assign an overall score. A perfect score is 10 points, with 9-10 points indicating high quality, nearly identical to the reference answer; 7-8 points indicating quality close to the reference answer; 6-7 points being of moderate quality; 4-5 points indicating a lower quality response; and 2-3 points for a response with significant errors.

Here is the true value response from the reference: {reference_input}

Here is the generated response from the assistant:
{predicted_input}

Here is the knowledge related to the workflow: 
{text format workflow knowledge}

Just reply with the score, the format is as follows,
Score: xxx
\end{lstlisting}

\paragraph{Task Completion.}
Below we provide the evaluation prompt for evaluating task completion, including the success rate and task progress.
\begin{lstlisting}[language={}, caption=Prompt for assessing task completion]
You serve as an assistant responsible for assessing if a dialogue system has achieved the user's goals. You are given the provided user profile, user objectives, and the dialogue record between the system and the user, your task is to determine if the system has met all the goals of the user.
Below is the user profile, the user's objectives, including the APIs the user expects to be called with the corresponding input parameters:

User target: {user target from the task-user summary}

Below is the workflow information (mermaid) and API information of the task where the dialogue is located.

workflow_info: {text format workflow knowledge}

Below is the interaction content between the role of 'user' and the 'assistant' system. In the assistant's 'Thought,' the content of 'Action' and 'Action_input' indicate the API and parameters that need to be called. The content of 'function' denotes the returned results of the API calls.

simulated_session: {the predicted session}

Now, your task is to decide whether the dialogue has fulfilled all the user's goals previously mentioned. This encompasses whether the dialogue has completed the information inquiry and interaction needed, if the corresponding APIs have been correctly called, and whether the API call parameters were transmitted accurately.

You only need to check whether the target within the provided workflow chart and API information has been completed. If the target goes beyond the scope of the provided workflow information, there is no need to check this target. You don't need to check whether the return value of the API parameter is reasonable, you only need to check whether the parameter collection is reasonable.

Only consider whether the system (assistant) has completed the task. If the user does not make any requests in the goal, or makes requests beyond the goal, the goal is considered completed. The main criteria are the collection and invocation of API parameters. You need to focus on examining whether the intent parsing and responses are reasonable, and whether each goal is completed. If all goals are completed, it is considered successful

You also need to output the total number of user goals (if there are multiple goals), and the number of goals that have been achieved, considering goals outside the business process as completed. (The total number of goals is greater than or equal to the number of completed goals).
Use a rigorous evaluation mode where each goal is achieved (or out of scope) for overall success

Your reply template should follow this format:
Result: yes/no (overall success)
Total number of goals: 4
Number of accomplished goals: 3
Reason: If the goal is completed, no reason is required. If not, please specify why you judged it to be incomplete.
\end{lstlisting}

\section{Additional Demonstration}\label{app:add_demo}
We further deliver a workflow knowledge showcase for `flight booking' within the role 'flight\_inquiry'  in the travel\&transportation domain, accompanied by a generated cross-scenario ground-truth session.

\paragraph{Text Format.}
Below is the workflow knowledge embedded in text format for the scenario of flight booking under the role of flight\_inquiry.
\begin{lstlisting}[language={}, caption=Text-Format workflow for flight booking]
The process of booking a flight begins when the user initiates the request. The user is then prompted to provide the flight ID. Once the flight ID is given, the system calls the checkAvailability function to confirm the flight status and check for availability based on the returned is_air status.

If the flight is available, the system informs the user about the flight's availability and requests their ID number and name. The system then calls the reserveFlight function to make the reservation and checks the success of the reservation based on the returned is_successful status.

If the reservation is successful, the system informs the user that the reservation succeeded and provides the booking details, which are based on other returned values from the reserveFlight. The user is then asked if they want to book another flight.

If the user decides to book another flight, the process loops back to requesting the user's ID number and name for the new reservation. If the user does not want to book another flight, they are welcomed to contact the system again for future needs.

If the flight is not available, the system informs the user that the flight is unavailable and asks if the user wants to book another flight. If the user decides to book another flight, the process returns to checking the availability of the new flight. If the user does not wish to book another flight, they are also welcomed to contact the system for future needs.

If the reservation fails, the system notifies the user of the failure and asks if they want to try booking another flight. If the user agrees, the process loops back to attempting the reservation again. If the user chooses not to rebook, they are informed that they are welcome to contact the system again in the future.

\end{lstlisting}

\paragraph{Code Format.}
Below is the workflow knowledge embedded in code format with python style for the scenario of flight booking.
\begin{lstlisting}[language=python, caption=Code-Format workflow for flight booking]
def book_flight():
    request_initiated = True

    while request_initiated:
        # Prompt for flight ID
        flight_id = request_flight_id()

        # Check availability
        if check_availability(flight_id):
            # Inform availability and request user info
            inform_user_available()
            user_id = request_user_id()
            user_name = request_user_name()

            # Attempt reservation
            reservation_result = reserve_flight(flight_id, user_id, user_name)
            if reservation_result['is_successful']:
                # Inform success and provide details
                inform_reservation_success(reservation_result)

                # Ask if user wants another booking
                if user_wants_to_book_another():
                    continue
                else:
                    request_initiated = False
                    inform_user_contact_again()
            else:
                # Inform failure and ask if user wants to try again
                inform_reservation_failure()
                if user_wants_to_try_again():
                    continue
                else:
                    request_initiated = False
                    inform_user_contact_again()
        else:
            # Inform unavailability and ask if user wants to book another flight
            inform_user_unavailable()
            if user_wants_to_book_another():
                continue
            else:
                request_initiated = False
                inform_user_contact_again()

def request_flight_id():
    return input("Please provide the flight ID: ")

def check_availability(flight_id):
    # Simulate check availability (always returns True for this example)
    return True

def inform_user_available():
    print("The flight is available.")

def request_user_id():
    return input("Please provide your ID number: ")

def request_user_name():
    return input("Please provide your full name: ")

def reserve_flight(flight_id, user_id, user_name):
    # Simulate reservation (always succeeds for this example)
    return {'is_successful': True, 'flight_details': 'Flight AA123 at 7:00 AM on April 5'}

def inform_reservation_success(reservation_result):
    print("Reservation succeeded.")
    print(f"Booking details: {reservation_result}")

def user_wants_to_book_another():
    response = input("Do you want to book another flight? (yes/no): ").strip().lower()
    return response == "yes"

def inform_user_contact_again():
    print("Thank you! Please contact us again for future needs.")

def inform_reservation_failure():
    print("Reservation failed.")

def user_wants_to_try_again():
    response = input("Do you want to try again? (yes/no): ").strip().lower()
    return response == "yes"

def inform_user_unavailable():
    print("The flight is unavailable.")

# Start booking process
book_flight()

\end{lstlisting}

\paragraph{Flowchart Format.}
Below is the workflow knowledge embedded in flowchart format for the scenario of flight booking. Its visualization is also shown in Figure~\ref{fig:app_visflowchart}.

\begin{lstlisting}[language={}, caption=Flowchart-Format workflow.]
flowchart TD
SK000(Start)--Book a flight-->SK001>Inquire the user for the Flight ID,
call checkAvailability to confirm flight status,
and check flight availability based on the returned is_air]
SK001--Flight is available--> SK002>Tell the user that the flight is available.
Ask the user for ID number and name, call reserveFlight, and check if the reservation is successful based on the returned is_successful ]
SK002--Reservation succeeded-->SK003[Inform the user that the reservation succeeded, and notify the user of booking details based on other returned values of reserveFlight. 
Ask if the user wants to book again]
SK002--Reservation failed-->SK004[Notify the user that the reservation failed and ask if the user wants to book again]
SK004  & SK003--User books again-->SK002
SK004 & SK003--User does not rebook-->SK006[The user is welcome to contact again for future needs]
SK001--Flight is unavailable-->SK005[Inform the user that the flight is unavailable, and confirm if the user wants to book another flight]
SK005--User books another flight-->SK001
\end{lstlisting}

\paragraph{Toolbox Information.}
Below we provide the tool information within the knowledge base of the flight booking scenario. There are two available tools of \textit{checkAvailability} and \textit{reserveFlight}.
\begin{lstlisting}[language={}, caption=Toolbox information example]
API: checkAvailability
API Desciption: Check the ticket availability for a flight given by the user
Input parameters: {"plan_code": {"type": "string", "description": "Flight ID, such as \"CA1234, CZ5678, MU9101, 3U4567, FM8901, HU7890\"", "required": true}, "estimated_time": {"type": "string", "description": "Estimated time, in the format: year + month + day + hour. For example, \"12 o'clock on January 7, 2199\". If not filled in, the default is now.", "required": false}}
Output parameters: {"is_air": {"type": "string", "description": "Whether the airplane can be ticketed. Enumerated type, [true, false]"}}

API: reserveFlight
API Desciption: Reserve the user's flight number and return the reservation information
Input parameters: {"plan_code": {"type": "string", "description": "Flight ID, such as \"CA1234, CZ5678, MU9101, 3U4567, FM8901, HU7890\"", "required": true}, "cabin_type": {"type": "string", "description": "Cabin type. Enumerated type, [Economy Class, Business Class, First Class]. If not filled in, these types will be selected from front to back", "required": false}, "estimated_time": {"type": "string", "description": "Estimated time, in the format: year + month + day + hour. For example, \"12 o'clock on January 7, 2199\". If not filled in, the default is now", "required": false}}
Output parameters: {"is_successful": {"type": "string", "description": "Whether the reservation is successful. Enumerated type, [true, false]"}, "boarding_number": {"type": "string", "description": "Boarding number, such as \"C1, D3, E7, F2, G4, H6\"."}, "seat_number": {"type": "string", "description": "Seat number, such as \"12A, 15C, 21F, 28D, 34B, 42E\""}, "airport_name": {"type": "string", "description": "Airport name, such as \"Beijing Capital International Airport, Shanghai Pudong International Airport\""}, "actual_time": {"type": "string", "description": "Actual time, in the format: year + month + day + hour. For example, \"12 o'clock on January 7, 2199\". If not filled in, the default is now."}}

\end{lstlisting}

\paragraph{Session Example}
Below we further provide a demonstration example of a cross-scenario ground-truth session for the role of flight\_inquiry, which includes the scenarios of flight booking and flight information search. As shown below, different interaction sources (user, agent, environment) are indicated upfront. The agent may choose to reply with a `Response' (interact with the user) or utilize a tool with `Action' along with `Action Input' (interact with the environment).

\begin{lstlisting}[language={}, caption=Ground-truth cross-scenario session example]
User:
Hello! I need to book a flight from New York to London for a business meeting. Can you provide options for flight times, duration, and pricing? Thank you!

Agent:
Response:  Could you please provide me with the departure date and time for your flight from New York to London? Additionally, if you have any preferred airlines, please let me know.

User:
Hi again! I'm looking to depart on April 5th, anytime in the morning. No specific airline preferences; I'm just looking for the best combination of price and travel time. Thanks!

Agent:
Response:  Could you specify an approximate time in the morning on April 5th for your departure? For example, would around 7 AM work for you?

User:
Absolutely, around 7 AM would be perfect. Could you also check if there are any flights available a bit later in the morning, just in case? Thanks!

Agent:
Action: planeSearch
Action Input: {"starting_point": "New York", "end_point": "London", "estimated_time": "7 o'clock on April 5, 2039"}

Environment:
{"plan_dict": ["plan_code": "AA123", "plan_type": " Boeing 737", "departure_time": April 5th, 2039, 7:00 AM", "plan_price": "$500"], ["plan_code": "BA456", "plan_type": " Boeing 777", "departure_time": April 5th, 2039, 8:00 AM", "plan_price": "$550"], ["plan_code": "DL789", "plan_type": "Boeing 787 ", "departure_time": April 5th, 2039, 9:00 AM", "plan_price": "$530"]}

Agent:
Response:  I found some flights for you from New York to London on the morning of April 5th. Here are the options:
1. Flight AA123: Departs at 7:00 AM, arrives at 7:00 PM, Duration: 12 hours, Price: $500
2. Flight BA456: Departs at 8:00 AM, arrives at 8:00 PM, Duration: 12 hours, Price: $550
3. Flight DL789: Departs at 9:00 AM, arrives at 9:00 PM, Duration: 12 hours, Price: $530
Please let me know which flight you would prefer or if you need more information on any of these options.

User:
Oh, thanks for the options! Could you tell me if any of these flights offer Wi-Fi service or extra legroom? I need to work during the flight.

Agent:
Response:  I apologize, but I don't have information regarding specific amenities like Wi-Fi service or extra legroom on flights. You might want to check directly with the airlines for detailed service offerings. Can I assist you with anything else related to your flight booking?

User:
That's understandable, thanks! Could you please help me book Flight AA123 then? I'll contact the airline directly for the extra details.

Agent:
Action: checkAvailability
Action Input: {"plan_code": "AA123", "estimated_time": "7 o'clock on April 5, 2039"}

Environment:
{"is_air":"true"}

Agent:
Response:  The flight AA123 on April 5th at 7:00 AM is available. To book this flight, could you please provide your ID number and full name as it appears on your identification?

User:
Sure! My full name is Johnathan L. Smith and my ID number is 987654321. Please proceed with the booking. Thanks for your help!

Agent:
Action: reserveFlight
Action Input: {"plan_code": "AA123", "estimated_time": "7 o'clock on April 5, 2039"}

Environment:
{"is_successful": "true", "boarding_number": "C1", "seat_number": "12A", "airport_name": "John F. Kennedy International Airport", "actual_time": "7 o'clock on April 5, 2039"}

Agent:
Response:  Your reservation for Flight AA123 on April 5th at 7:00 AM from John F. Kennedy International Airport has been successfully booked. Here are your booking details:
- Boarding number: C1
- Seat number: 12A
- Airport: John F. Kennedy International Airport
- Departure Time: 7 o'clock on April 5, 2039
Would you like to book another flight or is there anything else I can assist you with?

User:
No, that's all for now. Thanks for all your help!

Agent:
Response:  You're welcome, Johnathan! If you need any more assistance in the future, feel free to reach out. Have a great trip and a successful business meeting in London! Safe travels!

\end{lstlisting}

\end{document}